\documentclass{article}

\usepackage{microtype}
\usepackage{graphicx}
\usepackage{subcaption}
\usepackage{booktabs} % for professional tables

\usepackage{hyperref}
\usepackage{multirow}

\usepackage[preprint]{icml2026}

\usepackage{amsmath}
\usepackage{amssymb}
\usepackage{mathtools}
\usepackage{amsthm}
\usepackage{enumitem}

\usepackage[capitalize,noabbrev]{cleveref}

\theoremstyle{plain}

\theoremstyle{definition}

\theoremstyle{remark}

\usepackage{xspace}

\usepackage[textsize=tiny]{todonotes}
\usepackage{pifont}
\def\eg{\emph{e.g.,}\xspace} 

\def\ie{\emph{i.e.,}\xspace} 

\def\dataset{\textbf{Coper}\xspace}

\icmltitlerunning{Do Transformers Have the Ability for Periodicity Generalization?}
\begin{document}

\twocolumn[
  \icmltitle{Do Transformers Have the Ability for Periodicity Generalization?}
  \icmlsetsymbol{correspond}{*}
  % It is OKAY to include author information, even for blind submissions: the
  % style file will automatically remove it for you unless you've provided
  % the [accepted] option to the icml2026 package.

  % List of affiliations: The first argument should be a (short) identifier you
  % will use later to specify author affiliations Academic affiliations
  % should list Department, University, City, Region, Country Industry
  % affiliations should list Company, City, Region, Country

  % You can specify symbols, otherwise they are numbered in order. Ideally, you
  % should not use this facility. Affiliations will be numbered in order of
  % appearance and this is the preferred way.
% lhy, lige, dyh, sihanwu, peixu wang,  sihao cheng, taozhi chen, kechi zhang, hao zhu
  \begin{icmlauthorlist}
    \icmlauthor{Huanyu Liu}{pku}
    \icmlauthor{Ge Li}{pku,correspond}
    \icmlauthor{Yihong Dong}{pku}
    \icmlauthor{Sihan Wu}{pku}
    \icmlauthor{Peixu Wang}{pku}
    \icmlauthor{Sihao Cheng}{csh} \\
    \icmlauthor{Taozhi Chen}{thu}
    \icmlauthor{Kechi Zhang}{pku}
    \icmlauthor{Hao Zhu}{pku}
    \icmlauthor{Tongxuan Liu}{JD}
    %\icmlauthor{}{sch}
    %\icmlauthor{}{sch}
  \end{icmlauthorlist}
  \icmlaffiliation{pku}{School of Computer Science, Peking University}
  % \icmlaffiliation{CS}{}
  \icmlaffiliation{thu}{College of AI, Tsinghua University}
  \icmlaffiliation{csh}{School of Science and Engineering, The Chinese University of Hong Kong (Shenzhen)}
  \icmlaffiliation{JD}{jd.com}

  \icmlcorrespondingauthor{Huanyu Liu}{huanyuliu@pku.edu.cn}
  \icmlcorrespondingauthor{Ge Li}{lige@pku.edu.cn}
  \icmlcorrespondingauthor{Sihao Cheng}{sihaocheng@link.cuhk.edu.cn}
  % \icmlcorrespondingauthor{Firstname2 Lastname2}{first2.last2@www.uk}

  % You may provide any keywords that you find helpful for describing your
  % paper; these are used to populate the "keywords" metadata in the PDF but
  % will not be shown in the document
  \icmlkeywords{Machine Learning, ICML}

  \vskip 0.3in
]

% -------- hack: 手动无编号脚注 --------
\makeatletter
\renewcommand{\thefootnote}{*} % 脚注编号显示为 *
\footnotetext{Corresponding author.}
\renewcommand{\thefootnote}{\arabic{footnote}} % 改回阿拉伯数字
\makeatother
%\footnotetext[0]{Corresponding author.}
% this must go after the closing bracket ] following \twocolumn[ ...

% This command actually creates the footnote in the first column listing the
% affiliations and the copyright notice. The command takes one argument, which
% is text to display at the start of the footnote. The \icmlEqualContribution
% command is standard text for equal contribution. Remove it (just {}) if you
% do not need this facility.

% Use ONE of the following lines. DO NOT remove the command.
% If you have no special notice, KEEP empty braces:
\printAffiliationsAndNotice{}  % no special notice (required even if empty)
% Or, if applicable, use the standard equal contribution text:
% \printAffiliationsAndNotice{\icmlEqualContribution}

\begin{abstract}

Large language models (LLMs) based on the Transformer have demonstrated strong performance across diverse tasks. However, current models still exhibit substantial limitations in out-of-distribution (OOD) generalization compared with humans. We investigate this gap through periodicity, one of the basic OOD scenarios. Periodicity captures invariance amid variation. Periodicity generalization represents a model’s ability to extract periodic patterns from training data and generalize to OOD scenarios. We introduce a unified interpretation of periodicity from the perspective of abstract algebra and reasoning, including both single and composite periodicity, to explain why Transformers struggle to generalize periodicity. Then we construct \dataset about composite periodicity, a controllable generative benchmark with two OOD settings, \textbf{Hollow} and \textbf{Extrapolation}. Experiments reveal that periodicity generalization in Transformers is limited, where models can memorize periodic data during training, but cannot generalize to unseen composite periodicity. We release the source code to support future research\footnote{\url{https://github.com/gtxygyzb/periodicity_generalization}}.
% We further introduce the concept of composite periodicity to describe the combination and transformation of multiple periodicities.

\end{abstract}

%, motivating interest in their potential to achieve artificial general intelligence (AGI).

\section{Introduction}

\begin{figure}[ht]
    \centering
    \includegraphics[width=1.0\linewidth]{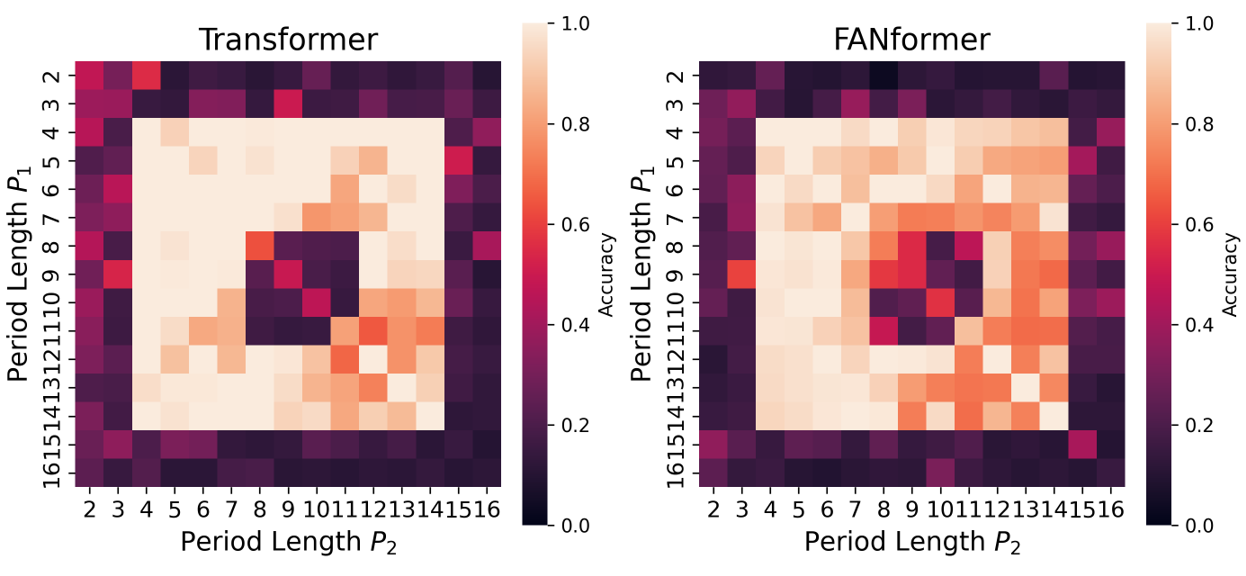}
    \vspace{-0.6cm}
    \caption{Accuracy heatmaps of Transformer and FANformer on composite periodicity tasks in \dataset. Models perform well on seen composite periods, but accuracy drops in Hollow and Extrapolation OOD settings, showing limits in periodicity generalization. The full results are shown in Figure~\ref{fig:RQ1_full_heatmaps}.}
    \vspace{-0.6cm}
    \label{fig:heatmap}
\end{figure}

% AGI需要OOD泛化 → 周期性是基本的OOD场景 → Transformer无法泛化composite periodicity → Transformer无法达到AGI

In recent years, the Transformer~\cite{transformer} has achieved milestone progress in modern artificial intelligence. Large language models (LLMs) based on the autoregressive architecture have demonstrated remarkable progress in natural language processing~\cite{GPT-4}, mathematical reasoning~\cite{deepseek_r1, openai_o3}, and code generation~\cite{codex, gemini-3}. Recent studies show that LLMs can even discover new physical laws~\cite{AI-newton} and prove mathematical theorems~\cite{AI_math}. These achievements lead many researchers to believe that LLMs may eventually reach artificial general intelligence (AGI)~\cite{Sparks_of_AGI}.

% LLMs such as OpenAI o3~\cite{openai_o3} and Google Gemini~\cite{gemini-3} surpass human performance in top-tier contests like the International Olympiad in Informatics (IOI).

% 近年来，Transformer 架构在人工智能领域取得了里程碑式的突破。基于 decoder-only 架构的大语言模型在数学推理与代码生成等任务上展现出惊人的能力；诸如 OpenAI o3、Google Gemini 等模型在 IOI、IMO 等国际竞赛任务上超越人类表现，能够证明数学定理、代码生成、推演形式化表达式、解决复杂推理问题。这些成果让不少研究者认为，基于Transformer架构的大语言模型，可能实现通用人工智能（AGI）。

However, some researchers remain skeptical about the path from LLMs to AGI. Compared with humans, who can learn complex patterns from limited data with low energy consumption~\cite{Ilya, How_Far_AGI}, current models still exhibit significant gaps in learning efficiency and out-of-distribution (OOD) generalization~\cite{Parrots, reasoning_failures, OOD_in_time_survey, Revisiting_Generalization, add_hollow}. For example, LeCun has argued that LLMs still have a fundamental gap in truly comprehending the complex world~\cite{LeCun}.

% 然而，也有部分人对LLM保持负面态度。与人类在有限样本与低能耗条件下即可学习复杂规律的能力相比，现有模型在\textbf{学习效率与域外泛化}方面仍存在显著差距。例如，Yann LeCun 对大语言模型的能力始终持怀疑态度，认为它们距离真正理解复杂世界的结构仍有根本差距

To understand this gap, periodicity provides a key entry point. Periodicity is an essential characteristic that appears widely in nature, mathematics, and human cognition~\cite{brain_rhythms, 1985book}. It reflects invariance amid variation, from nature~\cite{2020natural_rhythms} to abstract algebra~\cite{abstract_algebra}, and from signal processing~\cite{signal_processing} to reasoning~\cite{2017humanlearning, FAN}.
Periodicity generalization is the ability to extract periodic patterns from observed data and generalize to unseen scenarios.
For humans, it reflects the ability to perform abstract inference from known to unknown. For models, it reflects the ability to generalize learned periodic patterns from training data to OOD scenarios. % We think that OOD generalization is a necessary condition for achieving AGI, and periodicity is one of the basic OOD scenarios that exist in the natural world.

% 在理解这一差距时，\textbf{周期性（periodicity）}为这一讨论提供了关键切入点。周期性作为自然界与认知系统中普遍存在的结构特征，是自然界、数学以及认知系统中普遍存在的规律形式，从季节变化到群论结构，从信号处理到语言生成，它体现了变化中的不变性。对人类而言，周期性的建模本质上是基于已知对未知的一种抽象推断，是一种基于已知规律对未知的预测能力。对模型而言，对周期性的建模本质上是模型如何处理\textbf{域外数据}的能力。我们认为：\textbf{域外泛化是实现 AGI 的必要条件}。而周期性是自然界本身存在的基础域外数据结构形式之一。

Periodicity can be expressed using abstract algebra~\cite{abstract_algebra, dynamical_systems}. The essence of periodicity is invariance under specific transformations. When inputs or rules recur in temporal, spatial, or abstract forms, new instances can be inferred based on previously observed periodic patterns, which is a core feature of reasoning~\cite{FANformer}.

% Periodicity can be expressed using abstract algebra~\cite{abstract_algebra, dynamical_systems}. The essence of periodicity is invariance under specific transformations. At a deeper level, periodicity generalization reflects a system's (\ie human or LLMs) capability to imagine and reason about unseen scenarios. When inputs or rules recur in temporal, spatial, or abstract forms, the system can infer new instances based on previously learned periodicity, which is a core feature of reasoning~\cite{FANformer}.

% 周期性常可以在抽象代数的统一框架下得到系统表达。周期性的本质是“在某种变换下的不变性”。更深层地，周期性体现了系统对\textbf{未见结构的想象与推理能力}：当输入或规则在时间、空间或符号层面重复出现时，系统能够基于已学规律推演出新实例。这种规律抽象与跨分布外推能力，正是人类推理的核心特征之一。

To this end, we introduce a unified interpretation of periodicity from the perspective of abstract algebra and reasoning, describing periodicity in time-series and reasoning. Our interpretation explains why Transformers struggle to generalize periodicity. From this, we further introduce the concept of \textbf{composite periodicity}, which characterizes the combination and transformation of multiple periodicities.  %Our perspective provides new theoretical support for understanding model behavior in OOD scenarios.

% 基于此，本文从抽象代数与推理的视角，提出了一个统一的周期性解释（描述），用以描述时间序列中的周期性、推理中的周期性。在此基础上，我们进一步提出了\textbf{组合周期（composite periodicity）}的概念，用以描述多个周期叠加、组合与变换。该解释为理解模型在域外泛化场景下的表现提供了新的理论依据。

Then we construct \dataset, a dataset to characterize \underline{\textbf{Co}}mbi-nations and transformations of different \underline{\textbf{per}}iodicities, including training and test sets. The test set contains two challenging OOD settings:
\ding{182}~\textbf{Hollow}: the periods of the test samples remain within the training distribution, but certain period composition are deliberately removed from the training set, to evaluate the model's \textbf{interpolation ability}; %within the training distribution;
\ding{183}~\textbf{ Extrapolation}: the periods of the test samples lie beyond the training distribution, to evaluate the model's \textbf{extrapolation ability}.

% Then we构建了一个组合周期数据集，使不同的周期性产生叠加与变换，包括训练集和测试集。测试集中，我们设计了两类具有挑战性的设置： \textbf{（1）空洞（Hollow）}：测试样本的周期仍位于训练分布中，但我们刻意挖空训练集中的部分周期组合，使这些组合在训练中从未出现，从而衡量模型在训练分布内部的\textbf{插值能力}； \textbf{（2）外推（Extrapolation）}：测试样本的周期超出训练分布，用以衡量模型的\textbf{外推能力}。

As shown in Figure~\ref{fig:heatmap}, Experiments show that the Transformers can memorize seen composite periods during training, but fail in Hollow and Extrapolation OOD settings. Other model architectures (\eg Mamba, RWKV) and existing periodicity modeling works (\eg FANformer) exhibit similar failures. We futher discover three
interesting findings:
1) Transformers with RoPE fail to generalize single periodicity under non-invariant transformations.
2) Transformers struggle to generalize composite periodicity, even when the underlying single periods are seen during training.
3) Increasing data density or model scale improves performance in the Hollow setting, but only by a margin in extrapolation.

% Experimental results show that the Transformer can memorize periodic patterns seen during training in composite periodicity tasks. However, when faced with combinations never encountered in the training set, model performance drops significantly, and reasoning based on rule composition fails. Other model architectures (\eg Mamba, RWKV) and existing periodicity modeling works (\eg TFKAN, FANformer) exhibit similar failures.

% and discover three interesting findings

% We term this phenomenon \textbf{the failure of periodicity generalization}, indicating that models can memorize learned patterns but lack the ability to generate new inferences through composition, . 

% 实验结果表明，在组合周期任务中，Transformer 可以准确记忆和重现训练中出现的周期模式。但面对训练集中从未出现的周期组合时，模型性能显著下降，无法基于规则复用进行推理。我们将这种系统性的推理失效现象称为\textbf{周期性泛化失败（periodicity generalization）}：即模型能够记忆已学模式，但缺乏通过规则组合生成新推理的能力。类似地，其他模型架构（如 mamba、RWKV、KAN）以及现有周期性建模方法（如 FAN、fedformer）在该任务中也无法有效泛化。

The main contributions are summarized as follows:

\begin{enumerate}%[leftmargin=*,labelsep=0.5em,topsep=0pt,itemsep=0pt,parsep=0pt]
    \item \textbf{An interpretation of periodicity}: proposing a unified interpretation of single and composite periodicity based on group theory, to explain why Transformers struggle to generalize periodicity;
    \item \textbf{A dataset for composite periodicity}: constructing a controllable, generative \dataset dataset with Hollow and Extrapolation test scenarios. to evaluate models' periodicity generalization ability;
    \item \textbf{Limits of periodicity generalization}: experiments demonstrate that existing models can memorize seen periods, but fail to generalize through composition.
\end{enumerate}

%\begin{enumerate}
%    \item \textbf{抽象代数视角的周期性理论解释}：提出一个统一的周期性与组合周期性结构表征，使周期性可以在群论与不变性理论下形式化描述；
    
%    \item \textbf{提出组合周期推理基准数据集}：构建可控生成的组合周期任务，包含hollow he extrapolation两类测试场景，能够系统评估模型的泛化能力；

%    \item \textbf{揭示周期性泛化（periodicity generalization）失败现象}：实验证明现有模型虽能复用已见周期或模式，但在无法通过已学的reasoning pattern进行规则组合，从而泛化到新的组合周期。
%\end{enumerate}
\section{Related Work}
\label{sec:related_work}

\subsection{Out-of-distribution Generalization}
% 1. OOD指的是xxx
% 2. 一些工作讨论了OOD
Out-of-distribution (OOD) generalization studies how models generalize learned patterns when the test distribution differs from the training distribution~\cite{Towards_OOD,A_Survey_on_Evaluation}.
Several works~\cite{OOD_in_time_survey,WOODS} analyze OOD generalization of neural networks in time series and other real-world settings, showing that performance can drop sharply under distribution shifts, even when in-distribution accuracy is high.
Other works~\cite{OOD_algo_sudoku,Counting_Algo_Trans,Positional_Attention,A_Formal_Framework,Unlocking_OOD,Beyond_Single_Task,LM_Inductive_Count} investigate OOD generalization on logic and reasoning tasks, finding that standard Transformers can fit the training distribution but often fail on variations of the same underlying rules or on much longer inputs.

However, most existing works focus on a single task or a single family of rules at a time.
Generalization across composition of multiple tasks or rules, where models need to compose learned patterns, receives little study~\cite{Compositional_OOD_survey}.
Learning by Analogy~\cite{Learning_by_Analogy} introduces a causal framework for compositional generalization in visual reasoning.
Our work treats periodicity as a basic OOD scenario and provides a unified interpretation that links periodic patterns with rule composition.

% 3. 然而，并没有讨论多个任务的组合泛化性。
% 4. 《Learning by Analogy: A Causal Framework for Composition Generalization》在视觉领域研究了这个问题。我们旨在将周期性与规则进行统一的表述。

\subsection{Periodicity Modeling}
% 1. 一句话周期性很重要，是自然界中本身存在的xxx。
% 2. 一些工作尝试建模周期性，例如FedFormer 干了什么事。xxx干了什么事（列举3~4个）
Periodicity is a basic pattern that naturally appears in physical and biological processes. %and plays an important role in many forecasting and reasoning tasks~\cite{FAN}.
Several works~\cite{Autoformer,FEDformer,ETSformer,Periodformer,TFKAN} explicitly model periodicity in time-series forecasting to capture seasonal patterns and long-term dependencies. 
Autoformer~\cite{Autoformer} decomposes a time series into trend and seasonal components and uses an auto-correlation mechanism to capture periodicity for long-horizon forecasting.
FEDformer~\cite{FEDformer} introduces a frequency-enhanced Transformer that replaces standard attention with Fourier and wavelet blocks to improve long-horizon forecasting. 
More recently, ETSformer~\cite{ETSformer} combines exponential smoothing with attention and decomposes the series into level, growth, and seasonality components to produce more interpretable forecasts. %这个例子没位置可以删了

% More recently, TFKAN~\cite{TFKAN} employs a dual time–frequency KAN~\cite{KAN} architecture that models nonlinear relationships and periodicity in time and frequency domains.

% 3. 然而他们没有在OOD的场景上讨论
% 4. FAN和FANformer 从周期性与reasoning的关系出发，提出了傅里叶分析网络，训练了FANformer-1B。然而，在组合泛化任务上还是有所不足。
However, the above approaches only study periodicity in in-distribution settings and do not consider OOD scenarios. 
Fourier Analysis Network (FAN)~\cite{FAN} extends periodicity modeling to OOD scenarios using a Fourier-based module. 
FANformer~\cite{FANformer} further connects periodicity with reasoning and scales the FAN model to 1B parameters.
However, both still show limitations in composite periodicity tasks.

% In our composite periodicity tasks, both FAN and FANformer still show clear limitations when generalizing to unseen combinations of periodic rules.

%However, the above approaches are mainly evaluated on in-distribution forecasting benchmarks. 
%They do not analyze OOD generalization or compositional reasoning with periodic rules.
%Fourier Analysis Network (FAN)~\cite{FAN} connects periodicity with reasoning by introducing a Fourier-based module that explicitly encodes periodic functions.
%FANformer~\cite{FANformer} further integrates FAN into the attention blocks of large language models and scales FAN to a 1B-parameter model for efficient periodicity modeling.

% \input{section/definition}
\section{Unified Interpretation of Periodicity and Reasoning via Group Theory}
\label{sec:definition}

In this section, we present our unified interpretation of periodicity via group theory. Section~\ref{subsec:definition_p} formalizes the group-theoretic definition of periodicity. Section~\ref{subsec:rule_p} then generalizes the above definition to rule periodicity, grounded in concrete reasoning examples. 
Section~\ref{subsec:rope_p} explains the limitations of Transformers in generalizing rule periodicity.
Finally, Section~\ref{subsec:composite_p} introduces composite periodicity, extending this limitation to composed periodic rules.

%Section~\ref{subsec:rope_p} analyzes the ability of the Transformer to model periodicity.
%Finally, Section~\ref{subsec:composite_p} introduces the concept of composite periodicity.

% 在本节中，我们将our Unified Interpretation of Periodicity and Reasoning via Group Theory。在\label{subsec:periodicity}中，我们首先将从群论的视角定义周期性。接着，我们讲结合具体例子介绍推理中的规则周期性在\lable{subsec:regular_p}。基于此，我们在\lable{subsec:}分析了Transformer架构对周期性的建模能力。最后，我们在\lable 提出了组合周期性的概念。

\subsection{Group-theoretic Definition of Periodicity}
\label{subsec:definition_p}

Let $X$ be the value space (for example $X = \mathbb{R}$).
Consider a discrete sequence defined as a function $f : \mathbb{Z} \to X$.
Suppose that there exists $T \in \mathbb{N}^+$ such that
\[
f(t + T) = f(t), \quad \forall t \in \mathbb{Z},
\]
then we say $f$ admits the sequence period $T$.
Since $T \in \mathbb{N}^+$, we can assume that $T$ is the minimal positive period among all periods without loss of generality.
This description shows the periodicity in the temporal dimension.

From the perspective of abstract algebra, the essence of periodicity is to \textit{remain invariant under transformation} \cite{abstract_algebra, dynamical_systems}.
Let $X$ be an object space (for example, functions, numbers, or positions), and a group $G$ acts on $X$.
Suppose that there exists a non-identity element $g \in G$ and a minimal integer $n \in \mathbb{N}^+$ such that
\begin{equation}
g^n \cdot x = x,
\label{eq:group_p}
\end{equation}
then we say that the element $x \in X$ is \textbf{periodic} under the left $g$-action.
The minimal positive period is defined as follows:
\begin{equation}
T(x \mid g) = \min \left\{ n \in \mathbb{N}^+ \mid g^n \cdot x = x \right\}.
\label{eq:group_pT}
\end{equation}

The above definition naturally describes the invariance of periodic functions under group actions.
For instance, consider a periodic sequence:
\[
f : \mathbb{Z} \to \mathbb{R}, \quad f(t + T) = f(t), \ \forall t \in \mathbb{Z},
\]
the corresponding shift group is $G = \langle g \rangle$, and the left group action is defined to be\footnote{Here we use the inverse to ensure the associativity in the induced action. Otherwise, if $g \cdot f(t) := f(g \cdot t)$, then $g_1 \cdot (g_2 \cdot f(t)) = g_1 \cdot \phi_{g_2}(f)(t) = \phi_{g_2}(f)(g_1 \cdot t) = f(g_2 \cdot (g_1 \cdot t)) \neq f((g_1 \cdot g_2) \cdot t) = (g_1 \cdot g_2) \cdot f(t)$. The equation does not hold when the group is not abelian. Inverse solves this problem due to $(g_1 \cdot g_2)^{-1} = g_2^{-1} \cdot g_1^{-1}$.\label{fn:action}}
\[
g \cdot f \ := \ \phi_g(f) \ : \ t \mapsto f(g^{-1} \cdot t) = f(t - 1).
\]
Hence, the generator $g$ acts on functions repetitively:
\[
g^n \cdot f (t) = g^{n - 1} \cdot \phi_g(f) (t) = g^{n - 1} \cdot f (t - 1) = \cdots = f(t - n)
\]
Therefore, the minimal positive period under the group action on $f$ is
\[
\begin{aligned}
T(f \mid g) :&= \min \{ n \in \mathbb{N}^+ \mid g^n \cdot f = f \} \\
& = \min \{ n \in \mathbb{N}^+ \mid f(t - n) = f(t), \ \forall t \in \mathbb{Z}\} = T,
\end{aligned}
\]
We call this \textbf{sequence periodicity}, the repetitive pattern of the input sequence itself.

Our group-theoretic definition applies not only to sequence periodicity, but also to the characterization of rule periodicity (discussed in Section~\ref{subsec:rule_p}). When the mapping relationship is preserved for inputs or rules under transformations, it can also be viewed as invariance under the action of the corresponding transformation group.

% The abstract definition in this section provides a unified interpretation for the further discussion on regular invariance.

\subsection{Rule Periodicity in Reasoning}
\label{subsec:rule_p}

Reasoning tasks often exhibit \textbf{rule periodicity}, referring to the repetition of the same rule.
We still use the group action interpretation in Section~\ref{subsec:definition_p}.
Let the rule set be $\mathcal{R} = \{ R_k \}_{k \in I}$, the index set be $I = \mathbb{Z}$, and the value space be $V = \left\{ 0, \dots, p - 1 \right\}$, and we introduce two types of group:

\begin{itemize}
    \item \textbf{Shift group} $G_{\tau} = \langle \tau \rangle \simeq \mathbb{Z}$: \\
    the generator $\tau$ acts as $\tau \cdot k = k + 1$.
    \item \textbf{Modulo group} $G_{\sigma} = \langle \sigma \rangle \simeq \mathbb{Z}_p$: \\
    the generator $\sigma$ acts as $\sigma \cdot x = x + 1 \pmod p$.
\end{itemize}

% This section doesn't use f_1 and f_2
% Let the input be two sequences
% \[
% f_1(t) = (x_1,x_2,\dots,x_n),\qquad f_2(t) = (y_1,y_2,\dots,y_n),
% \]
Now we define two primitive reasoning rules as mappings (with position parameter $k$):

% \begin{itemize}

\paragraph{Digit-wise addition without carry:}
\[
R_k : \mathbb{Z} \times \mathbb{Z} \to \mathbb{Z}, \qquad
(x_k, y_k) \mapsto x_k + y_k.
\]
The shift group $G_{\tau} = \langle \tau \rangle$ acts on the index set $I$ as
$\tau \cdot k = k + 1$, and the induced action on the rule set $\mathcal{R}$ is\footnote{Here we use the inverse to ensure the associativity in the induced action. See footnote~\ref{fn:action}.}
\[
\tau \cdot R_k = R_{\tau^{-1} \cdot k} = R_{k - 1}, \quad \tau^n \cdot R_k = R_{k - n}.
\]
If there exists $T \in \mathbb{N}^+$ such that
\[
\tau^T \cdot R_k = R_k, \quad \forall k \in \mathbb{Z},
\]
then the rule $R_k$ is said to admit a period of length $T$ along the positional dimension. For digit-wise addition without carry\footnote{Digit-wise addition with carry can be expressed as a combination of the modulo group and the shift group; see Section~\ref{subsec:composite_p} for details on composite periodicity.}, the same rule is applied identically at adjacent positions. Therefore, $R_k$ satisfies $T = 1$.

The shift group $G_{\tau}$ acts on the rule set $\mathcal{R}$, rather than on a specific input sequence, which induces a high-dimensional group action over the entire function space.
As a concrete example, consider the computation of $342 + 117$:
\[
\begin{aligned}
\text{units: } & 2 + 7 = 9, \\
\text{tens: }  & 4 + 1 = 5, \\
\text{hundreds: } & 3 + 1 = 4,
\end{aligned}
\]
resulting in the sequence $459$, where the digit-wise addition rule is applied repeatedly at each position.

\paragraph{Single-position modulo:}
\[
M_k : \mathbb{Z} \to \mathbb{Z}_p, \qquad x_k \mapsto x_k \pmod p.
\]

The modulo group $G_{\sigma} = \langle \sigma \rangle$ acts on the value space $V$ as $\sigma \cdot x = x + 1 \pmod p$.
Under this action, the modulo mapping satisfies
\[
\begin{aligned}
M_k(\sigma^{T_2} \cdot x_k) & = (x_k + T_2) \pmod p \\
& = x_k \pmod p \\
& = M_k(x_k),
\end{aligned}
\]
so the modulo operation has the period length $T_2 = p$ on the value space.
For example:
\[
\begin{aligned}
& (8, 9, 7, 5) \bmod 4 \\
= \ & (8 \bmod 4,\, 9 \bmod 4,\, 7 \bmod 4,\, 5 \bmod 4) \\
\mapsto & (0,1,3,1),
\end{aligned}
\]
in which every module mapping on each position $x_k \mapsto x_k \pmod 4$ has the period length $T_2 = 4$ on $\mathbb{Z}_4$.

From the perspective of function space, the modulo operation can be viewed as the result of shift group acting on rule set $\mathcal{R} = \{ M_k : \mathbb{Z} \to \mathbb{Z}_p \}$,
\ie the same modulo rule being applied on different position $k$.
This shows the same rule periodicity as in the addition rules.
Here the period length for $M_k$ is $T_2' = 1$ on the index set.

In summary, digit-wise addition and single-position modulo operation can be seen as periodically invariant mapping, respectively under the action of \textbf{shift group} and \textbf{modulo group}.
This structural repetition shows the \textbf{rule periodicity} in reasoning: the same rule being repeated in position or value space. 
Rule periodicity reflects the model's capability of reusage and generalization.

\subsection{Limitation of Transformers in Periodicity Generalization}
\label{subsec:rope_p}

\subsubsection{Sequence Periodicity Captured by Transformers with RoPE}
\label{subsubsec:seq_periodicity}

In this section, we first clarify what kinds of periodicity that Transformers are able to generalize.
Due to the properties introduced by Rotary Position Embedding (RoPE), Transformers can generalize
some simple sequence periodicity, where periodic patterns are preserved through relative position information.
We formalize this capability below. Further analysis is provided in Appendix~\ref{app:B}.

\paragraph{Formal proof}
RoPE encoding can represent sequence periodicity: $e^{i \varphi(t)} = e^{i \theta t}, \ t \in \mathbb{Z}$ where $\theta = \frac{2\pi}{T}$.

% Emphasize the importance of phase difference
When we compute the cosine similarity of two embedding vectors $x_m, x_n$ with RoPE $e^{i \theta m}, e^{i \theta n}$, the position encoding turns out in a form of phase difference $\Delta \varphi_{m, n} = \varphi(m) - \varphi(n)$:
\[
\begin{aligned}
\vspace{-0.5cm}
(x_m \cdot e^{i \theta m}) \cdot (\overline{x_n \cdot e^{i \theta n}}) & = x_m \cdot \overline{x_n} \cdot e^{i \theta m} \cdot e^{-i \theta n} \\
& = x_m \cdot \overline{x_n} \cdot e^{i \theta (m - n)} \\
& = x_m \cdot \overline{x_n} \cdot e^{i \Delta \varphi_{m, n}}.
\vspace{-0.3cm}
\end{aligned}
\]
Because of this, RoPE focuses on the relative position rather than absolute position.
% phase difference is preserved by \varphi

So in essence, sequence periodicity of RoPE equals relative position invariance: if $f(t) = f(t + T), \ \forall t \in \mathbb{Z}$, then for all $a, b$, we have
\begin{equation}
    f(a) - f(b) = f(a + T) - f(b + T).
    \label{eq:3}
    \vspace{-0.3cm}
\end{equation}

RoPE preserves this relation by the phase difference $\Delta \varphi_{a, b}$. As a result, Transformers with RoPE can capture sequence periodicity satisfying Eq.~\eqref{eq:3}. Without RoPE (\eg using absolute positional encodings), the periodicity generalization ability of Transformers is substantially weaker. Experimental evidence supporting our proof is provided in the top of Figure~\ref{fig:sec3} (see Appendix~\ref{app:RoPE_SinPE} for details).

\begin{figure}
    \centering
    \includegraphics[width=1.0\linewidth]{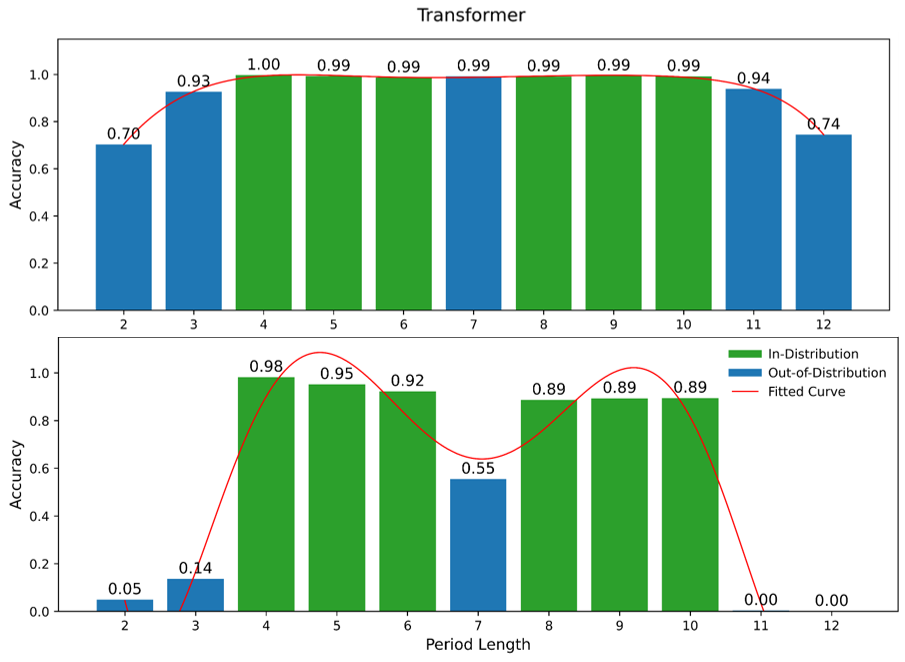}
    \caption{Top: Transformer with RoPE generalizes well for transformations satisfying translation invariance ($f(t+T)=f(t)$). Bottom: Single-period sequences under a non-invariant transformation ($f(t+T)=2\cdot f(t)$), showing the generalization failure.}
    %\vspace{-0.3cm}
    \label{fig:sec3}
\end{figure}

% such as sinusoidal positional encoding

%Transformer 能通过ROPE实现部分满足eq3的周期性泛化
%如果没有ROPE，Transformer的周期泛化能力会更差，见附录B.2
%%%%%%%%%%%%%%%%%%%%%

% reasoning task关键。

% limitation：只证明了ROPE给transformer带来的先验知识，transformer架构的其他组件主要是通过实验。

\subsubsection{Limitation in Capturing Rule Periodicity}
\label{subsubsec:rule_periodicity}

In reasoning tasks, \textbf{rule periodicity} is more crucial, \ie repeating and reusing the same rule on different structures.
For instance, digit-wise addition and modulo operation both satisfy:
\[
R_{a + T} = R_a.
\]

Rule periodicity is different from sequence periodicity: a rule $R_k$ is applied to sequence $f$ and obtains $R_k(f(a))$, and it repeats on every position \textbf{without relying on relative position difference} of the sequence. Now we present a concise proof that Transformer fails to represent such rule periodicity.

\paragraph{Formal proof} Assume $R_k$ possesses a period that is different from $T$, which is the period of $f$.
On one hand, $f$ satisfies $g^T \cdot f = f$, where $\langle g \rangle$ is the shift group acting on the function space.
On the other hand, $R_k$ does not exhibit the same shift invariance.
In general, the two group actions do not commute with $R_k$, \ie 
\begin{equation}
\vspace{-0.1cm}
    \tau^T \cdot R_k \neq R_k \circ g^{-T}.
    \label{eq:group_no_equal}
% \vspace{-0.1cm}
\end{equation}
If Eq.~\eqref{eq:group_no_equal} holds, we would have
\[
\begin{aligned}
    \vspace{-0.1cm}
    & \ R_{a + T}(f(a + T)) - R_{b + T}(f(b + T)) \\
    = & \ R_{a + T} (g^{-T} \cdot f(a)) - R_{b + T} (g^{-T} \cdot f(b)) \\
    = & \ \tau^{-T} \cdot R_{a + T}(f(a)) - \tau^{-T} \cdot R_{b + T}(f(b)) \\
    = & \ R_a(f(a)) - R_b(f(b)).
    \vspace{-0.1cm}
\end{aligned}
\]
Unfortunately, Eq.~\eqref{eq:group_no_equal} does not hold.
So for all $a, b \in \mathbb{Z}$, generally we have
\begin{equation}
    \begin{aligned}
    \vspace{-0.2cm}
    & R_a(f(a)) - R_b(f(b)) \\
    \neq \ & R_{a + T}(f(a + T)) - R_{b + T}(f(b + T)),
    \vspace{-0.2cm}
    \end{aligned}
    \label{eq:R_no_equal}
\end{equation}
which means that rule periodicity cannot be represented by RoPE’s relative position information.

\paragraph{Numerical counterexample}
Let RoPE's implicit period $T = 4$, the encoding $e^{i \varphi(t)}$ with phase $\varphi(t) = \frac{2\pi}{T}t$.
The rule be $\alpha(a) = a \pmod 3$ (with obvious period $T' = 3$).

Consider $a = 0, b = 1$. The relative position information $\Delta \varphi_{a, b}$ satisfies:
\[
\begin{aligned}
\vspace{-0.2cm}
& \varphi(0) - \varphi(1) = \varphi(a) - \varphi(b) \\
= \ & \varphi(8) - \varphi(9) = \varphi(a + 2T) - \varphi(b + 2T).
\vspace{-0.2cm}
\end{aligned}
\]

But:
\[
\begin{aligned}
& \alpha(0) - \alpha(1) = 0 - 1 = -1 \\
\neq \ & \alpha(8) - \alpha(9) = 2 - 0 = 2,
\end{aligned}
\]
\ie rule periodicity doesn't satisfy relative position invariance, and there doesn't exist a function $f$ rely solely on RoPE phase $\varphi(t)$ to determine the rule $R(t)$.

Hence, Transformer with RoPE can represent \emph{sequence periodicity} satisfying Eq.~\eqref{eq:3}, but cannot represent \emph{rule periodicity} according to Eq.~\eqref{eq:R_no_equal}. Experimental evidence supporting our proof is provided in the bottom of Figure~\ref{fig:sec3} (see Appendix~\ref{app:disable_ROPE} for details).

% transformer利用ROPE是不能泛化eq5所表示的规则周期性的，见附录B.3实验。

%in Section~\ref{subsec:composite_p}.

%To avoid the interference of other factors in Transformer's architecture, in further experiment we fix the Tokenizer and Embedding\footnote{we adopt the Tokenizer and Embedding of Qwen2.5 and freeze them}, to exclude extra parameter bias in the learning process.
%Furthermore, this article proves that even if equipped with shift invariance prior via RoPE, Transformer architecture still cannot represent more complex rule.

\subsection{Composite Periodicity}
\label{subsec:composite_p}
To further explore the limitation of Transformer architecture, we introduce the formal definition of \textbf{composite periodicity}. Assume there are two periodic sequences:
\[
f_1 : X_1 \to Y_1, \quad f_2 : X_2 \to Y_2,
\]
and define composite operation:
\begin{equation}
C(f_1(t), f_2(t)) := (M \circ R)(f_1(t), f_2(t)),
\label{eq:composition}
\end{equation}
for example modular addition:
\[
\vspace{-0.3cm}
C(f_1(t), f_2(t)) = (f_1(t) + f_2(t)) \pmod p
\]

We further define the group structure of \textbf{basic reasoning modulo rule}.
Let $G_R = G_{\tau} = \langle \tau \rangle$ be the shift group acting on addition rules in position space, and $G_M = G_{\sigma} = \langle \sigma \rangle$ be the modulo group acting on the modulo operation rules.
In the perspective of group theory, the composite group acting on reasoning modulo rules can be presented by direct product:
\begin{equation}
    G = G_R \times G_M = \langle \tau \rangle \times \langle \sigma \rangle,
    \label{eq:composite_G}
\end{equation}
and the induced action is
\begin{equation}
    \begin{aligned}
    (\tau^i, \sigma^j) \cdot (R_k, M_k) & = (R_{\tau^{-i} \cdot k}, M_k \circ \sigma^j) \\
    & = (R_{k - i}, M_k(\cdot + j)).
    \end{aligned}
    \label{eq:composite_induced_action}
    \vspace{-0.2cm}
\end{equation}

If the composite group $G$ defined by Eq.~\eqref{eq:composite_G} is \textbf{non-invariant} according to Eq.~\eqref{eq:R_no_equal}, Transformers cannot generalize such composite periodicity. 

Next, Section~\ref{sec:dataset} constructs datasets based on our interpretation, and Section~\ref{sec:experiment} evaluates the performance of multiple models on various composite periodicity tasks as support.

% Section~\ref{sec:dataset} \& \ref{sec:experiment} empirically evaluate the Transformer's performance in representing complex rules on composite periodicity.
\section{Dataset}
\label{sec:dataset}

To investigate the generalization ability of existing model architectures in composite periodicity scenarios, we construct \dataset, a fully controllable and generative dataset for composite periodicity. \dataset contains 50k training samples and 3k test samples.
Using our data construction script (see supplementary material), \dataset can be easily scaled to construct more samples.
Moreover, by modifying our script, new composite periodicity tasks (\eg the convolution) can be constructed.

% and can be easily scaled to generate more samples and additional periodicity composition rules according to experimental requirements (see Section~\ref{sec:experiment}).

We compose the rules $R_k$ and $M_k$ introduced in Section~\ref{subsec:composite_p}, and define a modulo addition operator $C$, which is explicitly unfolded along the time dimension as
%为了探究现有模型架构在组合周期场景下的泛化能力，我们构建了A dataset for composite periodicity: constructing a controllable, generative dataset. 我们选取Section~\ref{subsec:rule_p}中举例的规则$R_k$和$M_k$进行组合，定义带模加法操作$C$, which在时间轴上显式展开为：
\[
\begin{aligned}
C(f_1(t), f_2(t))
&= (M \circ R)(f_1(t), f_2(t)) \\
&= \bigl( f_1(t \bmod P_1) + f_2(t \bmod P_2) \bigr) \bmod P .
\end{aligned}
\]
For example, let $f_1(t) = (1, 2, 3)$ with period $P_1 = 3$ and $f_2(t) = (1, 2)$ with period $P_2 = 2$. Then, their composition via Eq.~\eqref{eq:composition} produces
\[
\begin{aligned}
C(f_1(t), f_2(t)) &= (f_1(t \bmod 3) + f_2(t \bmod 2)) \bmod 10 \\
     &= (2, 4, 4, 3, 3, 5, \dots).
\end{aligned}
\]
If the periodicity conditions in Eqs.~\eqref{eq:group_p} \&~\eqref{eq:group_pT} are satisfied, such periodicity compositions can also be interpreted under our proposed unified interpretation (\eg convolution tasks; see Section~\ref{subsec:exp_composition} for experiments).

We set the training period range as $L \le P_1, P_2 \le U,$
where $L$ and $U$ denote the lower and upper bounds of the training range, respectively. The training set contains the combinations
\[
\mathcal{P}_{train} = \{ (P_1,P_2) \mid P_1,P_2 \in [L,U], (P_1,P_2) \notin \mathcal{P}_{test}^{hollow} \}.
\vspace{-0.1cm}
\]

During dataset construction, we deliberately introduce two types of OOD settings, \textbf{Hollow} and \textbf{Extrapolation}, to analyze two types of model generalization: interpolation and extrapolation.
Each OOD setting contains 1k test samples, in addition to 1k in-distribution test samples. More hyperparameters are listed in Appendix~\ref{app:exp_details}.

\begin{figure}[t]
    \centering
    \includegraphics[width=0.95\linewidth]{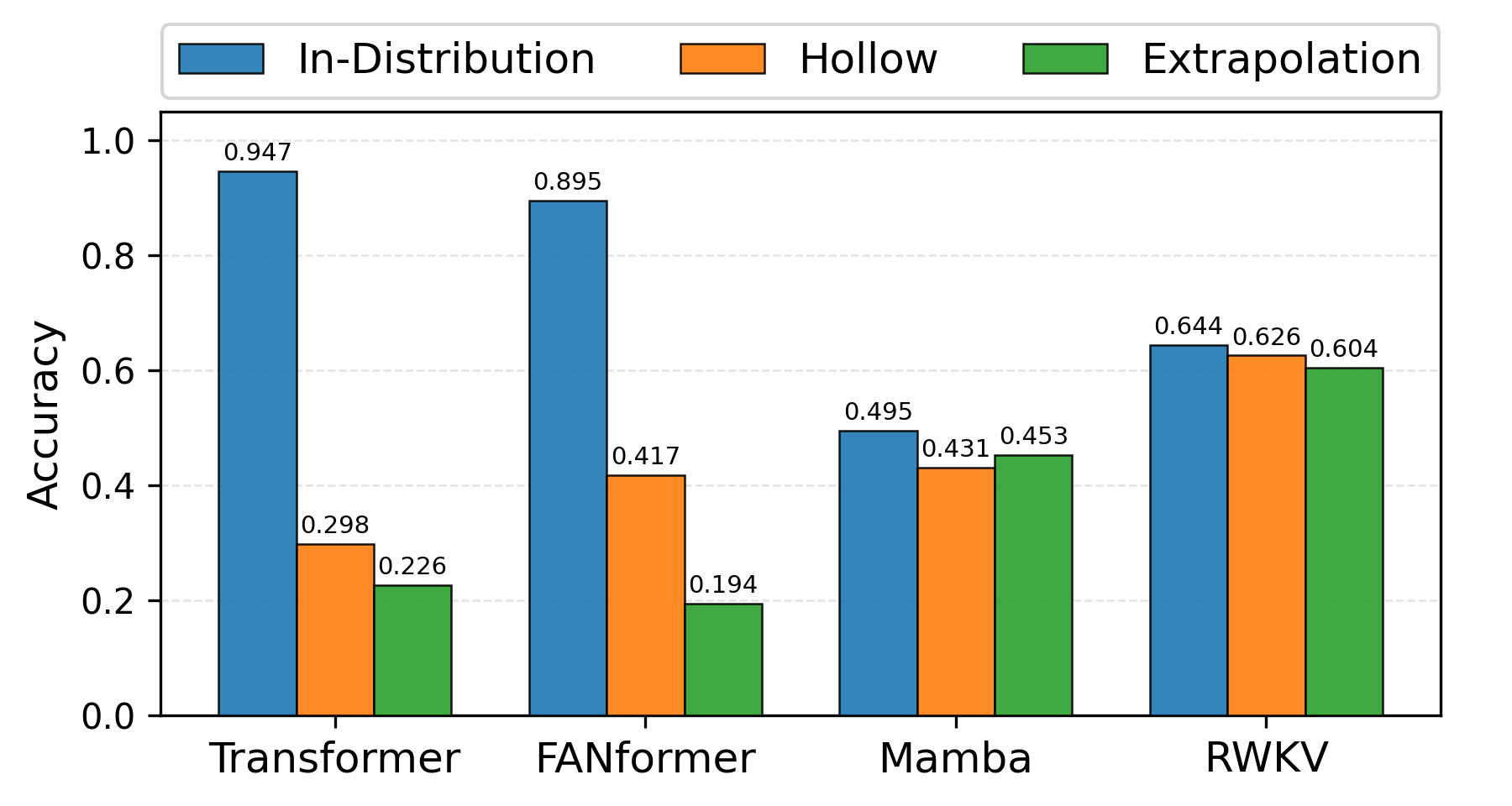}
    % \vspace{-0.2cm}
    \caption{Accuracy by category on composite periodicity tasks.}
    \vspace{-0.3cm}
    \label{fig:category_accuracy}
\end{figure}

\paragraph{(1) Hollow}

We define the deliberately held-out hollow combinations within the training range as
\[
\mathcal{P}_{test}^{hollow} = \{ H_1, H_2, \dots, H_m \} \subset [L,U]^2,
\]
These hollows are located inside the training range $[L,U]$, adjacent to trained combinations; however, they are not included in training.

For example, when $[L,U] = [3,10]$ and $m=2$ hollow combinations are selected:
\[
H_1 = (5,6), \: H_2 = (6,7) \quad \ie \; \mathcal{P}_{test}^{hollow} = \{ H_1, H_2\}.
\]
Hollow evaluates the model's interpolation ability to infer unseen composite periodic rules from adjacent, in-distribution combinations.

\paragraph{(2) Extrapolation}

We define the new combinations outside the training range as
\[
\mathcal{P}_{test}^{extra} = \{ (P_1,P_2) \mid 
P_1 \notin [L,U] \;\text{or}\; P_2 \notin [L,U] \}.
\]

For example, when $[L,U] = [3,10]$, extrapolation combinations can be: $(2,11),\ (11,12),\ (2,12)$.
Extrapolation evaluates the model's extrapolation ability to infer OOD composite periodic rules from the training distribution.

% 如果能满足\ref{eq:group_p}与\ref{eq:group_pT}的周期性，更多的rule composition也可以套用群理论（见Section~\ref{subsec:exp_composition}更多的例子）

% 在构建数据集时，我们刻意引入两类域外情形：\textbf{Hollow}与 \textbf{extrapolation}（训练域外扩展），以分析模型的两种泛化能力——插值与外推。
\section{Experiments}
\label{sec:experiment}

\begin{figure*}[t]
    \centering
    \includegraphics[width=1.0\linewidth]{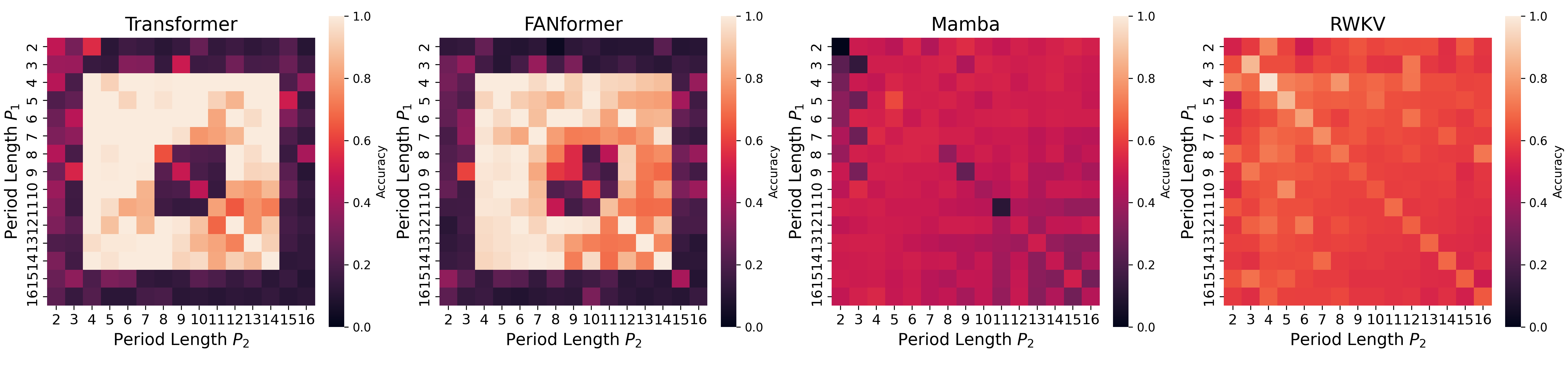}
    \caption{Full accuracy heatmaps across all baseline models on composite periodicity tasks, corresponding to the final models trained as in Figure~\ref{fig:loss_curve}. Notably, even when trained for 1000 epochs, RWKV and Mamba still fail to fit.}
    % \vspace{-0.2cm}
    \label{fig:RQ1_full_heatmaps}
\end{figure*}

\begin{figure}[t]
    \centering
    \includegraphics[width=0.9\linewidth]{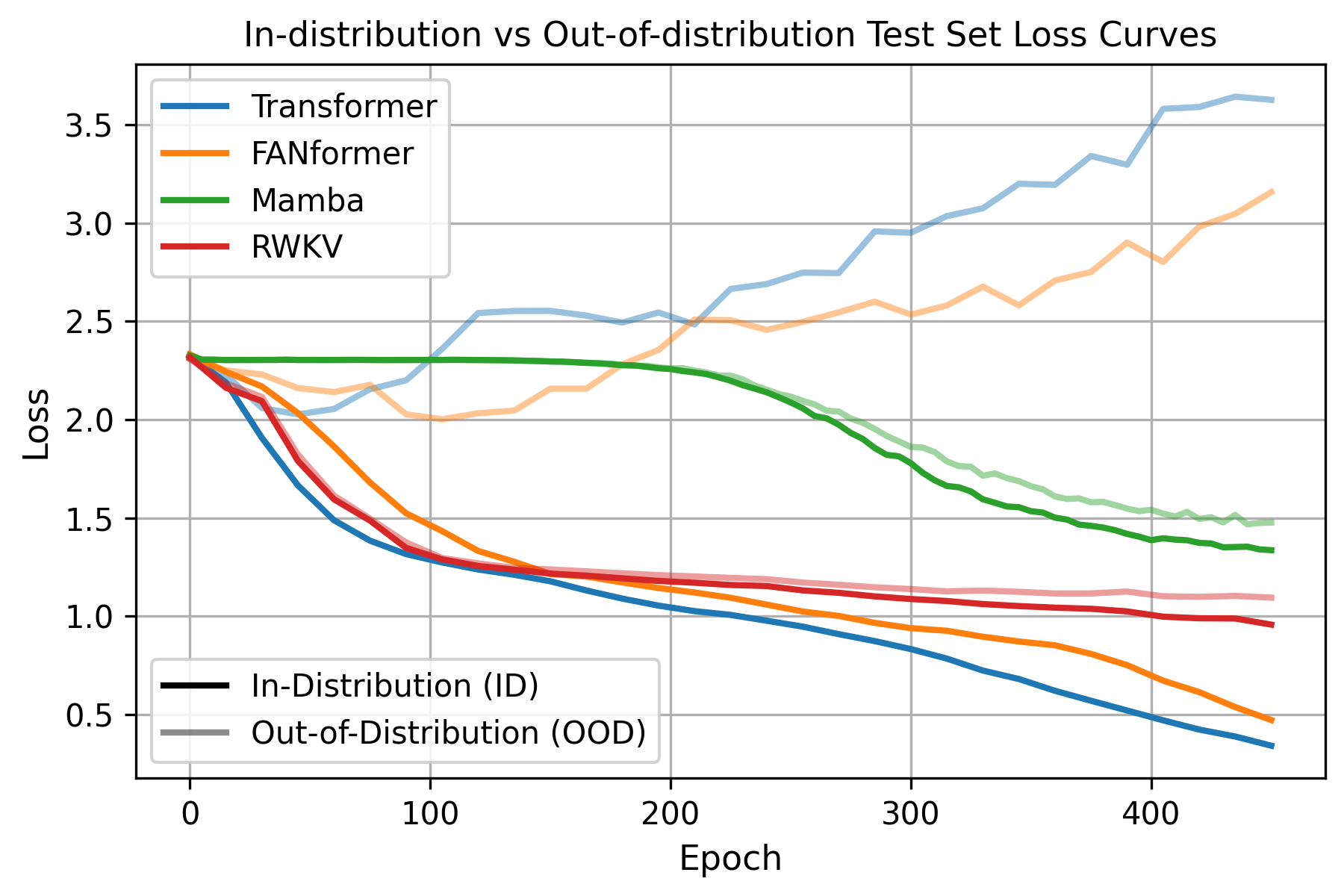}
    % \vspace{-0.1cm}
    \caption{Loss curves on ID and OOD testsets during training.}
    % \vspace{-0.3cm}
    \label{fig:loss_curve}
\end{figure}

\subsection{Failure of Periodicity Generalization Across Different Architectures}
\label{subsec:exp_architecture}

%  在第一个实验中，我们探究不同模型组合周期性的拟合与泛化能力。除了Transformer，我们还选用了目前周期性建模代表性的架构：FANFormer、KAN，以及近期讨论度较高的架构：Mamba、RWKV作为baseline。Unless otherwise specified, all models are trained and evaluated using the same random seeds and hyperparameter settings (provided in Appendix). 训练曲线图如fig:loss_curve所示。For evaluation metrics, 我们使用测试集id和ood的loss，以及测试集in-distribution accuracy,  Hollow and Extrapolation三类的准确率， 并报告中总平均值。实验结果如表tab:arch_comparison、图fig:category_accuracy fig:heatmap所示

In the first experiment, we evaluate the fitting and generalization abilities of different model architectures about composite periodicity. In addition to the Transformer, we include representative architectures for periodicity including FANFormer~\cite{FANformer}, as well as recently discussed architectures, Mamba and RWKV as baselines. And different models' loss curves on in-distribution and OOD test sets during training are shown in Figure~\ref{fig:loss_curve}. Unless otherwise specified, all models are trained and evaluated using the same random seeds and hyperparameter settings (provided in Appendix~\ref{app:exp_details}). For evaluation metrics, we report the final test loss on the in-distribution and OOD splits, test accuracy on the in-distribution, Hollow, and Extrapolation splits, together with the overall average. The results are reported in Table~\ref{tab:arch_comparison}, Figure~\ref{fig:RQ1_full_heatmaps} and Figure\ref{fig:category_accuracy}. 

\begin{table}[t]
\centering
\caption{Performance comparison across different model architectures on composite periodicity tasks. ID denotes in-distribution.}
\setlength{\tabcolsep}{4pt}
\resizebox{1.0\linewidth}{!}{
\begin{tabular}{lcc|cccc}
\toprule
\multirow{2}{*}{Model}
& \multicolumn{2}{c|}{Testset Loss $\downarrow$} 
& \multirow{2}{*}{ID} 
& \multirow{2}{*}{Hollow} 
& \multirow{2}{*}{Extrapolation} 
& \multirow{2}{*}{Avg.} \\
\cmidrule(lr){2-3}
& ID 
& OOD
& 
& 
& 
& \\
\midrule
Transformer & 0.34 & 3.63 & 94.7 & 29.8 & 22.6 & 49.0 \\
Fanformer   & 0.47 & 3.15 & 89.5 & 41.7 & 19.4 & 50.2 \\
% TFKAN         & 0.45 & 3.73 & 91.5 & 31.7 & 20.9 & 48.0 \\
% \midrule
Mamba       & 1.33 & 1.48 & 49.5 & 43.1 & 45.3 & 46.0 \\
RWKV        & 0.95 & 1.09 & 64.4 & 62.6 & 60.4 & 62.5 \\
\bottomrule
\end{tabular}
}
\label{tab:arch_comparison}
\vspace{-0.3cm}
\end{table}

\begin{figure}[ht]
    \centering
    \includegraphics[width=1.0\linewidth]{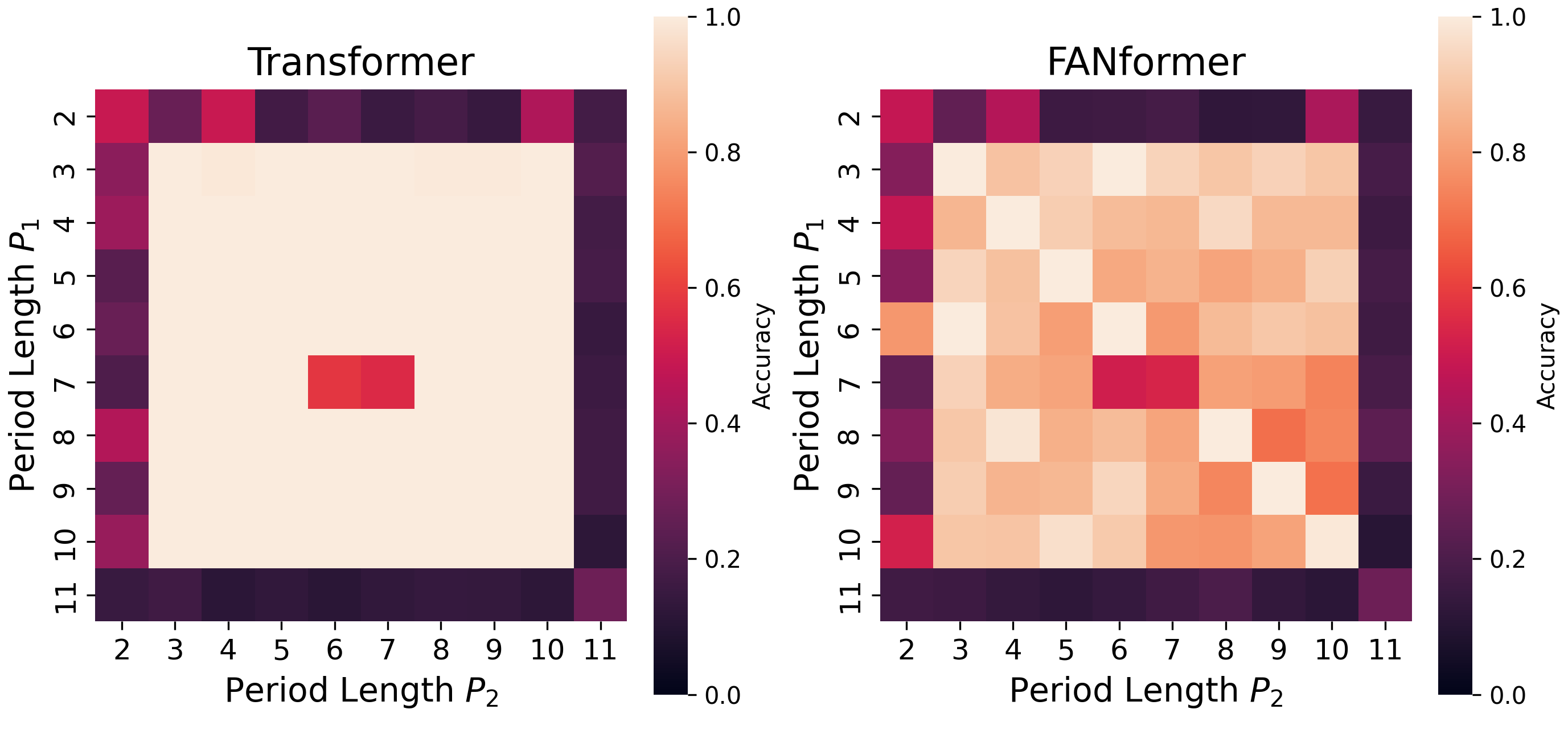}
    % \vspace{-0.3cm}
    \caption{Accuracy heatmaps with increased training data density.}
    % \vspace{-0.3cm}
    \label{fig:rq2_heatmap}
\end{figure}

The results show that the Transformer fails to generalize composite periodicity in OOD scenarios. Although the Transformer fits the training distribution well, with 94.7\% in-distribution accuracy, its accuracy on unseen combinations is low, with only 29.8\% on Hollow and 22.6\% on Extrapolation.
Architectures designed for periodicity modeling do not resolve this limitation. FANFormer also achieves high in-distribution accuracy at 89.5\%, while performance on Hollow and Extrapolation remains low at 41.7\% and 19.4\%.
In contrast, state-based models show much weaker fitting under the same training budget. Importantly, OOD generalization can only be interpreted when in-distribution fitting is sufficient. RWKV reaches 62.6\% on Hollow and 60.4\% on Extrapolation, but in-distribution accuracy is only 64.4\%, far below the Transformer. Overall, existing architectures fail to support both accurate in-distribution fitting and OOD generalization simultaneously, and none truly learn the composite periodicity $R$.

\begin{figure*}[t]
    \centering
    \includegraphics[width=1.0\linewidth]{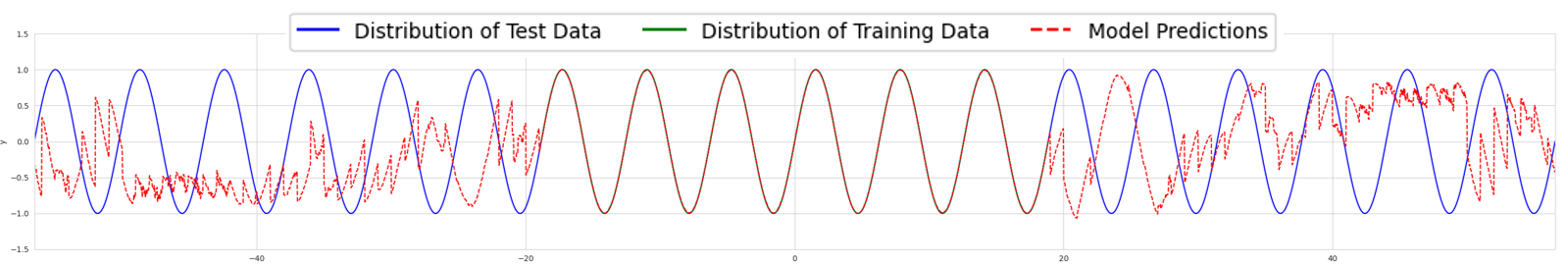}
    \vspace{-0.5cm}
    \caption{Transformer fails to generalize $y = \sin(x)$, where x is a token sequence.}
    \vspace{-0.2cm}
    \label{fig:trans_sin}
\end{figure*}

\subsection{Effect of Data Density OOD Generalization}
\label{subsec:exp_density}

In the second experiment, we further investigate the effect of training data density on OOD generalization. Intuitively, the Hollow lies within the training distribution, and the neighboring period combinations of the Hollow are all observed during training. A natural question is whether reducing the hollow size, \ie making the training data denser, enables models to fill the missing combinations via interpolation. To this end, while keeping the number of training samples fixed, we shrink the period range from $[2, 16]$ to a denser range of $[2, 11]$ to explicitly control the hollow size, retaining only a hollow combination $H = (6, 7)\cup(7, 7)$. 

The results are shown in Figure~\ref{fig:rq2_heatmap} and Figure\ref{fig:rq2_category}. Increasing training data density improves interpolation in the Hollow setting. For example, the Transformer achieves 56.5\% on Hollow, compared with 29.8\% reported in Table~\ref{tab:arch_comparison}, indicating an increase of +26.7\% through denser training data. However, performance on Extrapolation remains largely unchanged (22.7\% vs 22.6\%). FANFormer exhibit similar trends: higher accuracy in Hollow but minimal margin in Extrapolation. These findings indicate that denser training data improves performance in the Hollow, but has minimal margin on Extrapolation, showing that models do not learn the composite periodicity $R$. More details are shown in Appendix \ref{app:RQ2_full}.

% 在本节中，我们进一步探究训练数据密度对 Hollow 场景下模型插值能力的影响。直观来看，Hollow 设置中的测试组合位于训练分布内部，相邻的周期组合在训练集中均已出现，因此一个自然的问题是：当空洞规模减小、训练数据在周期组合空间中更加稠密时，模型是否能够通过插值填补空洞，从而缓解 Hollow 场景下的性能退化？为此，我们在保持训练样本数量不变的前提下，缩短周期范围，显式控制 Hollow 的规模，仅保留单个最小空洞组合H1=(5, 6). thereby alleviating the performance degradation in the Hollow setting

% 增加数据密度能提升OOD generation，但对外推提升有限。 在hollow transformer达到了56.5%%，相比起Table~\ref{tab:arch_comparison}的29.8%提高了xxx，但是在外推场景22.7%与22.6%仍然无提升。FANformer，KAN也有相似的结论。这表明模型能通过插值学会空洞，但是无法外推。

\subsection{Scaling Model and Data Size for Generalization}
\label{subsec:exp_scaling}

In the third experiment, we study the effect of model scale on generalization of composite periodicity. The previous experiments show that, under a fixed model scale, both different architectures and different data densities suffer from clear generalization failures in the Hollow and Extrapolation settings. A natural question is whether these failures are caused by insufficient model capacity. We gradually increase the number of layers to scale model parameters and evaluate the generalization performance in the Hollow and Extrapolation settings. All models are trained for the same number of epochs on identical training data. %The results are reported in Table~\ref{tab:scaling_comparison}.

% 第三个实验旨在探究模型参数规模对组合周期性泛化能力的影响。前述实验表明，在固定模型规模下，不同架构、不同数据量在 Hollow 与 Extrapolation 场景中均表现出显著的泛化失败。一个自然的问题是，这种失败是否源于模型容量不足，抑或是架构本身在组合规则建模上的结构性局限。为此，我们在保持模型架构与训练设置一致的前提下，通过逐步增大模型深度（层数）来扩展参数规模，系统分析模型规模提升是否能够改善其在 Hollow 与 Extrapolation 场景下的泛化表现。实验结果如\ref{tab:scaling_comparison}所示。

The results of scaling model parameters are reported in Table~\ref{tab:scaling_comparison}. Increasing the number of layers consistently improves generalization in both Hollow and Extrapolation settings. For instance, the Transformer improves from 29.8\% to 57.2\% on Hollow and from 22.6\% to 32.7\% on Extrapolation when scaling from 3 to 7 layers. FANFormer show similar trends, with OOD loss decreasing from 3.21 to 2.09. These results show that larger models improve OOD generalization for composite periodicity. However, Extrapolation performance remains below in-distribution accuracy, and such improvement comes at the cost of computational overhead, which will be discussed in Section~\ref{sec:discussion}. More details are shown in Appendix \ref{app:RQ3_full}.

\subsection{Failure under Other Composite Periodicity}
\label{subsec:exp_composition}

Previous experiments analyzed the failure of generalization for composite periodicity from the perspectives of model architecture, data density, and model scale. However, they are based on the same composite periodicity. A key question is whether the observed failure depends on the specific rule periodicity or represents a more general phenomenon, as suggested in Section~\ref{subsec:rope_p}. Prior work conducted experiments on carry-addition tasks~\cite{add_hollow}. In our final experiment, we use the \textbf{circular convolution} task as a new composite periodicity rule to further evaluate OOD generalization. The group-theoretic interpretation of circular convolution is provided in Appendix \ref{app:circular_convolution}.

\begin{table}[t]
\centering
\caption{Performance comparison across different model architectures on circular convolution.}
\setlength{\tabcolsep}{4pt}
\resizebox{1.0\linewidth}{!}{
\begin{tabular}{lcc|cccc}
\toprule
\multirow{2}{*}{Model}
& \multicolumn{2}{c|}{Testset Loss $\downarrow$} 
& \multirow{2}{*}{ID} 
& \multirow{2}{*}{Hollow} 
& \multirow{2}{*}{Extrapolation} 
& \multirow{2}{*}{Avg.} \\
\cmidrule(lr){2-3}
& ID 
& OOD
& 
& 
& 
& \\
\midrule
Transformer & 0.42 & 3.16 & 80.6 & 15.9 & 17.8 & 38.1 \\
Fanformer   & 1.24 & 1.66 & 47.4 & 34.6 & 24.9 & 35.6 \\
%TFKAN       & 0.48 & 3.33 & 78.1 & 14.2 & 17.3 & 36.5 \\
%\midrule
Mamba       & 1.99 & 2.05 & 17.7 & 15.4 & 17.2 & 16.8 \\
RWKV        & 2.11 & 2.15 & 14.2 & 11.9 & 12.1 & 12.7 \\
\bottomrule
\end{tabular}
}
\vspace{-0.5cm}
\label{tab:RQ4_conv}
\end{table}

The results are reported in Table~\ref{tab:RQ4_conv}. Under circular convolution, models exhibit the same generalization failure observed in previous experiments. The Transformer achieves 80.6\% in-distribution accuracy but performs poorly on Hollow and Extrapolation, with accuracies of 15.9\% and 17.8\%, respectively. %FANformer shows a similar pattern, reaching 78.1\% in-distribution accuracy while failing on Hollow and Extrapolation.
State-based models like Mamba and RWKV fail to fit the training distribution within the same training budget, reflecting low learning efficiency. Overall, the generalization failure persists across more composite periodicity scenarios, rather than being specific to a unique case. More details are shown in Appendix~\ref{app:RQ4_full} and Appendix~\ref{app:more_composite}.

% 前述实验已经从模型架构、数据密度和模型参数规模等方面系统分析了组合周期性泛化失败的现象。然而，这些分析均基于同一类组合周期规则。一个关键问题是，该泛化失败是否依赖于具体规则形式，抑或是Section 3.3.2中证明的一种更普遍的现象。已有工作主要讨论了进位加法场景下的 Hollow 现象。在最后一个实验中，我们选用循环卷积任务作为新的测试规则，以进一步验证模型在不同组合周期性规则下的泛化表现。循环卷积的群论解释见附录。

% 模型：transformer、fanformer、TF-KAN、（Mamba RWKV/RNN）

% 实验一、周期性泛化失败现象，放一个热力图，其他的丢附录。所有模型的训练Loss图，以及三类测试集的柱状图放一起。

% 实验二、数据密度/空洞大小的影响（例如只挖一个空）

% 实验三、scaling law的表格：同一个模型，不同参数量、不同的performance

% 实验四、不同的组合规则、卷积、更多的模型架构，等规则组合形式。

\section{Discussion}
\label{sec:discussion}

\subsection{More Types of Periodicity}

Beyond sequence and rule periodicity, there exist \textbf{hidden periodicity} patterns, where individual samples do not show obvious repetition, and the periodicity only appears at the level of the full data distribution.
An example is the sine function.
Rather than being contained in one piece of data (\eg $12341234 \cdots$), the periodic patterns in the sine function can be learned only after seeing many the $(x, y)$ data points across the distribution.
We conduct an experiment to train a Transformer to fit $y = \sin(x)$.
Consistent with \dataset, we divide the data into in-distribution $[-3\pi, +3\pi]$ for training and OOD beyond this range.
As shown in Figure~\ref{fig:trans_sin}, the Transformer fails to generalize such periodicity.
More experimental details are provided in Appendix~\ref{app:hidden_periodicity}. In real-world tasks, periodicity appears as combinations of different types.

%In practical tasks (\eg natural language processing), periodicity patterns appear in complex forms and usually intertwine with each other.
%This requires the models to possess the capability to tackle basic periodicity first.

\subsection{Hollow, Data Density, and Scaling Behavior}

OOD performance is influenced by data density and model size, which helps explain ``grokking" behavior observed during scaling~\cite{grokking}. Larger models and denser data enable stronger generalization beyond the training distribution.
However, our group-theoretic interpretation and experiments show that increasing data density or model size does not enable rule generalization. Even with minimal Hollow size, missing periodic combinations remain unresolved. As large-scale pretraining data approaches saturation, further improvements require exponentially higher training cost or model parameters. Progress beyond Hollow and Extrapolation therefore requires model architectures to have at least the ability to represent reasonable OOD generalization, like periodicity. 

% Under this setting, models answer unseen but similar natural language inputs.
%Even if the training distribution matches the full human knowledge distribution, the model cannot achieve intelligence beyond the human or extend the boundaries of human knowledge. 

%\subsection{Out-of-Distribution Generalization and AGI}

%Recent progress of Transformer models has renewed interest in AGI. However, simple and deterministic tasks, such as character counting, remain difficult. These tasks can also be decomposed into simple composite periodicity, which represents only one type of OOD generalization achievable through learning $R$.
%Our experiments use fully controllable composite periodicity tasks without noise or semantic ambiguity, where Transformer models fail in both Hollow and Extrapolation scenarios. The failure reflects the absence of rule reuse and rule composition. Model outputs follow interpolation or extrapolation induced by the training distribution, rather than compositional reasoning. When inference falls outside the training distribution, models generate incorrect predictions. 
%The same mechanism may explain why models make mistakes on simple tasks.
%Such results indicate that scaling the current Transformer architecture and training paradigm does not support reliable OOD generalization, which is required for AGI, even artificial superintelligence (ASI).
\section{Conclusion and Future Work}
\label{sec:conclusion}

Our work proposes a unified interpretation of periodicity and reasoning through group theory to explain why Transformers struggle to generalize periodicity. Based on this, we further propose the composite periodicity dataset \dataset to enable controlled evaluation of periodicity generalization. Experiments show that current models fail in Hollow and Extrapolation scenarios, revealing the limitations of current architectures in periodicity generalization.

Our limitations include 1) not studying external reasoning mechanisms such as chain-of-thought (CoT) or memory, and 2) applying our interpretation only to Transformer with RoPE, without more components like the self-attention. Future works include 1) extending the group-theoretic interpretation to more reasoning scenarios and 2) designing architectures that can model composite periodicity.

% 我们提出了一个Unified Interpretation of Periodicity and Reasoning via Group Theory。基于此理论，提出了组合周期数据集\dataset。实验发现xxx

% 我们工作的limitation是，只讨论了模型架构本身，并未讨论模型外推理的一些技术例如CoT、memory对OOD泛化性的影。在未来，我们会：1. 将群论的解释推广到更多场景 2. 设计能解决OOD组合周期性的模型。

% Acknowledgements should only appear in the accepted version.

\section*{Impact Statement}

This paper presents work whose goal is to advance the field of Machine
Learning. There are many potential societal consequences of our work, none
which we feel must be specifically highlighted here.

\bibliography{ref}
\bibliographystyle{icml2026}

%%%%%%%%%%%%%%%%%%%%%%%%%%%%%%%%%%%%%%%%%%%%%%%%%%%%%%%%%%%%%%%%%%%%%%%%%%%%%%%
%%%%%%%%%%%%%%%%%%%%%%%%%%%%%%%%%%%%%%%%%%%%%%%%%%%%%%%%%%%%%%%%%%%%%%%%%%%%%%%
% APPENDIX
%%%%%%%%%%%%%%%%%%%%%%%%%%%%%%%%%%%%%%%%%%%%%%%%%%%%%%%%%%%%%%%%%%%%%%%%%%%%%%%
%%%%%%%%%%%%%%%%%%%%%%%%%%%%%%%%%%%%%%%%%%%%%%%%%%%%%%%%%%%%%%%%%%%%%%%%%%%%%%%
\newpage
\appendix
\onecolumn
\section{Experimental Details of Training and Dataset}
\label{app:exp_details}

\begin{table}[h!]
\centering
\caption{Training hyperparameters and dataset parameters.}
\resizebox{0.9\textwidth}{!}{
\begin{tabular}{l c l c}
\toprule
\multicolumn{2}{c}{\textbf{Training Hyperparameters}} & \multicolumn{2}{c}{\textbf{Training Hyperparameters}} \\
\midrule
Batch size & 32 & Number of epochs & 450 \\
Learning rate & $1 \times 10^{-5}$ & Weight decay & 0.01 \\
Hidden state size & 896 & & \\
\midrule
\multicolumn{2}{c}{\textbf{Dataset Parameters}} & \multicolumn{2}{c}{\textbf{Dataset Parameters}} \\
\midrule
Number of training samples & 50{,}000 & Number of test samples & 3{,}000 \\
Sequence period range (train) & $[4,14]$ & Total range & $[2,16]$ \\
In-distribution pairs & $(P_1,P_2) \in [4,14]^2 \setminus \text{Hollow}$ & Hollow test pairs & $P_1,P_2 \in [8,11]$ \\
Extrapolation test pairs & $(P_1 \text{ or } P_2 \notin [4,14])$ & Output token & $0\sim 9$ \\
\bottomrule
\end{tabular}}
\label{tab:training_params_2seq_add}
\end{table}

\paragraph{Dataset Construction.} Training sequences are sampled such that
\[
(P_1,P_2) \in [4,14]^2 \setminus \text{Hollow},
\]
where \textbf{Hollow} defines a held-out region within the training range for interpolation evaluation.  

The test set contains two types of OOD scenarios:

\begin{itemize}
    \item \textbf{Hollow}: combinations within $[4,14]^2$ but excluded from training, \ie $P_1,P_2 \in [8,11]$.
    \item \textbf{Extrapolation}: combinations where $P_1$ or $P_2$ is outside the training range $[4,14]$, \eg $(1,16),(2,17)$, to evaluate extrapolation ability.
\end{itemize}

Each sequence is generated by sampling two sequences of length $P_1$ and $P_2$, repeating them to the least common multiple length, summing element-wise modulo 10, and concatenating as \texttt{sequence1 + '+' + sequence2 + '='}. The next-token prediction task is then defined over this sequence.

We choose to use the pretrained Qwen2.5 tokenizer and embedding and keep them fixed during training to avoid introducing spurious numeric priors. Our prior experiments show that training embeddings from scratch on numeric tasks inadvertently allows the model to capture numeric magnitude information directly, which can lead to shortcut solutions and hack the intended evaluation of reasoning capability. By using fixed pretrained embeddings, we ensure that models rely on learned reasoning rather than trivial numeric cues, enabling a fair comparison across all models.

We train the model three times using fixed random seeds, evaluate each run separately, and report the averaged results. And for each model, we compute accuracy by performing token-wise comparisons between the model’s predicted sequence and the ground-truth sequence, and averaging the correctness over all positions and all samples in the dataset.

%To avoid the influence of embeddings, we use the Qwen2.5 tokenizer and embedding and keep them fixed during training to ensure fair comparison across all models.

%%%%%%%%%%%%%%%%%%%%%%%%%%

\section{Further Analysis: Capacity and Limitations of Transformer with RoPE in Periodicity Modeling}
\label{app:B}

\subsection{Formal Illustration of RoPE and Single Period Shift Invariance}
\label{app:rope_def_p}

In Section~\ref{subsec:rope_p}, we dive into rule periodicity. Here we analyze how Transformer models \textbf{simple sequence periodicity} satisfying Eq.~\eqref{eq:3} through RoPE, for example the time shift invariance of a single periodic sequence $f(t) = f(t + T)$.

The core idea of RoPE is to map time step $t$ onto the phase rotation in the complex plane:
\[
\mathrm{RoPE}(x_t) = x_t e^{i \theta t}, \quad \theta = \frac{2\pi}{T},
\]
where $T$ is the implicit period length.
When a time shift $\Delta t$ occurs on the input sequence, we have
\[
\frac{\mathrm{RoPE}(x_{t + \Delta t})}{\mathrm{RoPE}(x_t)} = \frac{x_{t + \Delta t} \cdot e^{i \theta (t + \Delta t)}}{x_t \cdot e^{i \theta t}} = \frac{x_{t + \Delta t}}{x_t} \cdot e^{i \theta \Delta t},
\]
and if the periodicity ensures $x_{t + \Delta t} = x_t$, then
\[
\mathrm{RoPE}(x_{t+\Delta t}) = e^{i \theta \Delta t} \cdot \mathrm{RoPE}(x_t),
\]
hence shift equivariance remains in the self-attention computation:
\[
\langle \mathrm{RoPE}(x_{t+\Delta t}), \mathrm{RoPE}(x_{s+\Delta t}) \rangle
= \langle \mathrm{RoPE}(x_t), \mathrm{RoPE}(x_s) \rangle.
\]
This means that RoPE can realize the sequence shift invariance under single period $T$, therefore captures the simple \textbf{sequence periodic pattern} of the input sequence. We conduct an experiment to compare the ability of modeling single periodicity between RoPE and traditional absolute positional encoding (\ie Sinusoidal Positional Encoding, SinPE) in Appendix~\ref{app:RoPE_SinPE}.

\begin{figure}[h!]
    \centering
    \includegraphics[width=0.7\linewidth]{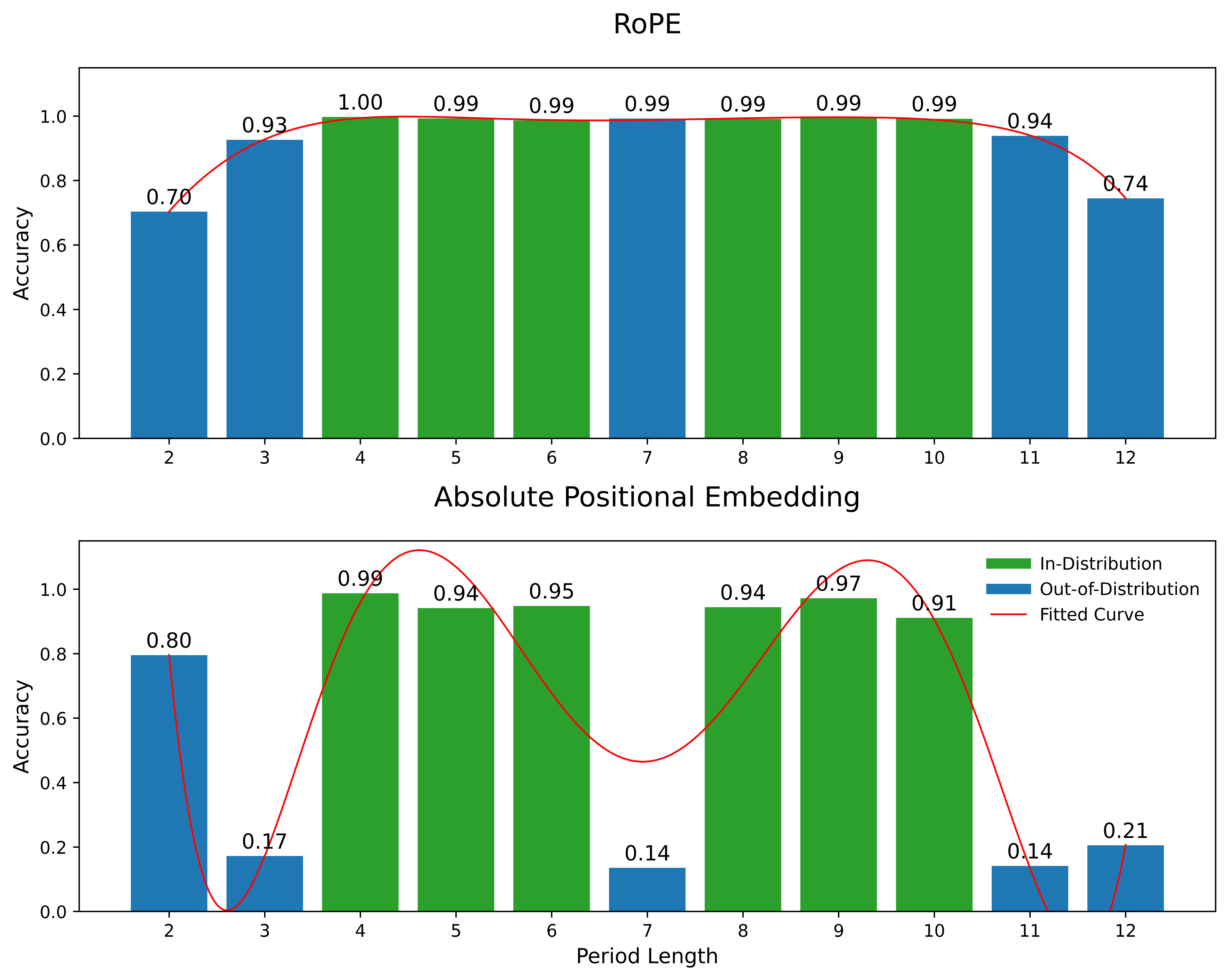}
    \caption{Comparison of periodicity generalization between RoPE and absolute positional encoding.}
    \vspace{-0.5cm}
    \label{fig:RoPE_SinPE}
\end{figure}
\begin{figure}[h!]
    \centering
    \includegraphics[width=0.55\linewidth]{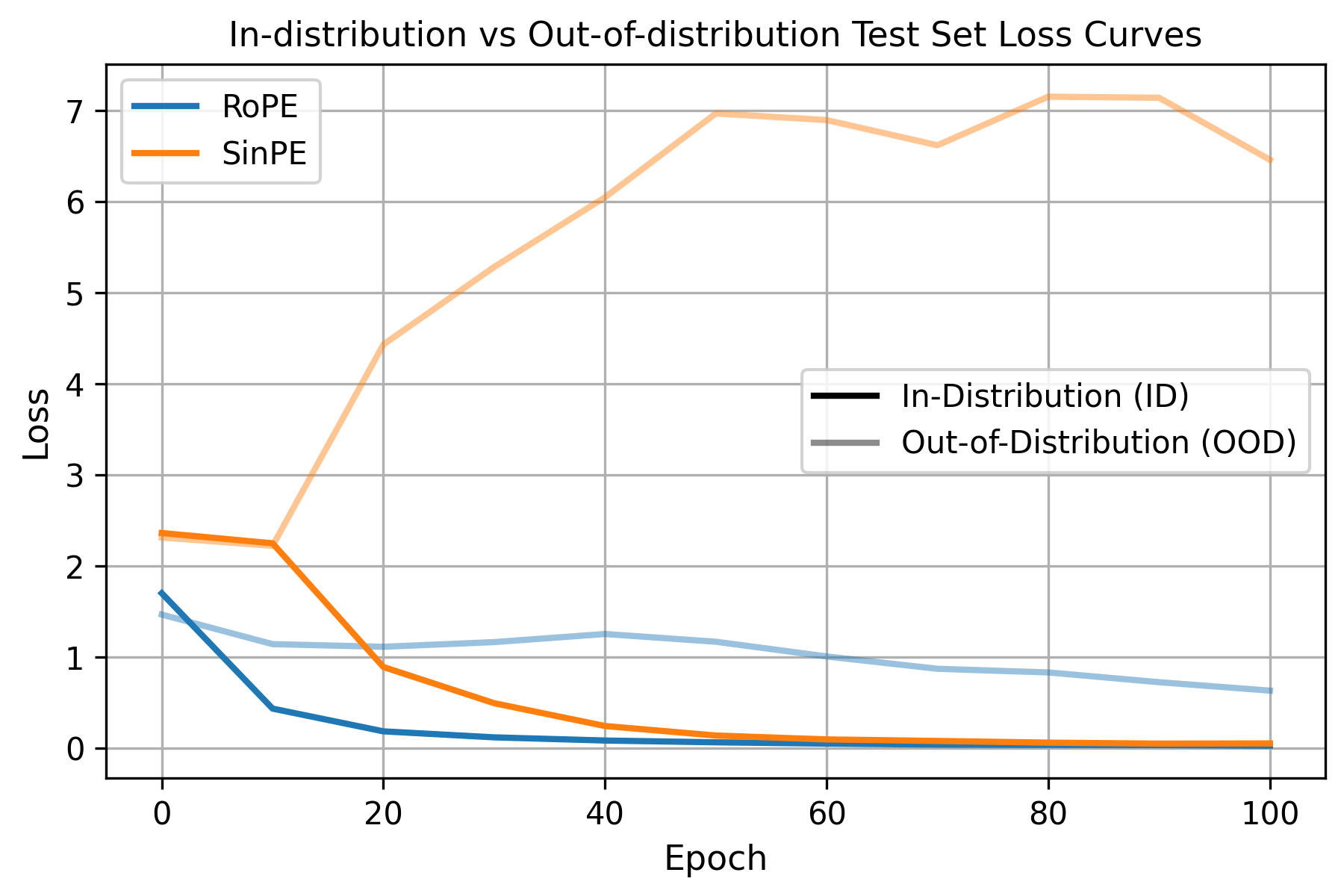}
    \caption{Loss curves for models with RoPE and absolute positional encoding on single-periodicity generalization tasks. The SinPE means Sinusoidal Positional Encoding, which is an absolute positional encoding in the original Transformer.}

    \label{fig:single_yes_loss_curve}
    
\end{figure}

\subsection{Experimental Validation of RoPE Modeling Single Periodicity}
\label{app:RoPE_SinPE}

In Appendix~\ref{app:rope_def_p}, we analyze how Transformer models simple sequence periodicity through RoPE. To prove that RoPE contributes to this, we conduct an experiment to compare RoPE with absolute positional encoding on the task of generalizing single periodicity.

We construct a periodic sequence continuation task. Given a periodic numerical sequence with more than two cycles, the model is required to predict the subsequent tokens. For example, here is an example for $T = 6$, where a single cycle is \texttt{955884}:
\[
\text{Input:} \quad \texttt{9558849558849} \qquad \text{Target output:} \quad \texttt{55884955884}
\]
The datasets are constructed as follows:
\begin{itemize}
    \item \textbf{Training Set (In-distribution)}: Sequences with periods $T \in \{ 4, 5, 6, 8, 9, 10 \}$, totaling 10,000 samples.
    \item \textbf{Test Set (OOD)}: Totaling 3,000 samples.
    \begin{itemize}
        \item \textbf{Hollow}: Sequences with periods $T = 7$.
        \item \textbf{Extrapolation}: Sequences with periods $T \in \{ 2, 3, 11, 12 \}$.
    \end{itemize}
\end{itemize}
We compare two transformer architectures: one with RoPE and another with absolute positional encoding. Both models are trained until convergence (100 epochs). The accuracy we use to measure the performance is at the token level, which is the ratio of correctly predicted tokens across all predicted ones.

The result is shown in Fig.~\ref{fig:RoPE_SinPE}, where the two graphs represent the performance of two positional encoding, and the two colors stand for in-distribution and OOD generalization ability. Fig.~\ref{fig:single_no_loss} reports loss of both positional encodings on in-distribution and OOD datasets during training. We draw the following conclusions:

In the in-distribution scenario ($T \in \{ 4, 5, 6, 8, 9, 10 \}$), RoPE and absolute positional encoding both achieve near 1.00 accuracy, indicating their common ability to memorize the seen periodic patterns.
In the case of hollow interpolation ($T = 7$), RoPE remains a high accuracy, but the performance of absolute positional encoding drops significantly. In extrapolation cases ($T \in \{ 2, 3, 11, 12 \}$), the accuracy of RoPE declines at a certain degree. On the contrast, the accuracy of absolute positional encoding is far less than RoPE's. The failure of absolute positional encoding on generalizing OOD cases proves its incompetence in modeling single periodicity.
Notably, when $T = 2$, absolute positional encoding gains accuracy $\approx 0.80$, which is due to the features of absolute positional encoding. It captures low-range parity information, and gains advantages through even period lengths in training sets, which is not an evidence to the extrapolation ability of absolute positional encoding.

The result confirms the theoretical conclusion in Appendix~\ref{app:rope_def_p}. By utilizing relative positional invariance compared to absolute positional encoding, RoPE has the ability to model single periodic patterns and make effective generalization. However, although RoPE performs well in generalizing single periodicity, it fails to generalize when facing composite periodicity as in Section~\ref{subsubsec:rule_periodicity}.

\subsection{The Failure of RoPE Modeling Single Periodicity under Non-invariant Transformation}

\label{app:disable_ROPE}

\begin{figure}[ht]
    \centering
    \includegraphics[width=0.7\linewidth]{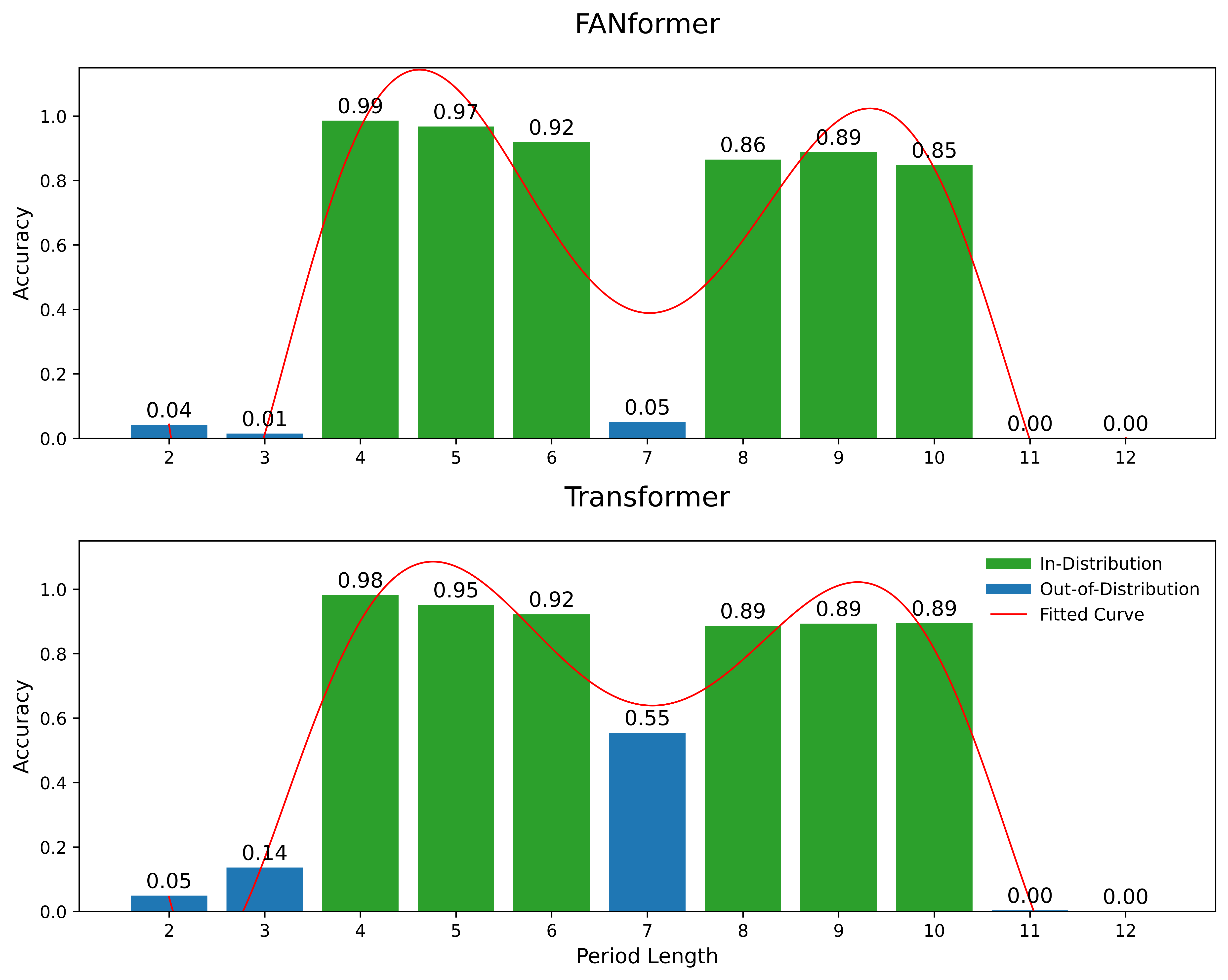}
    \caption{Prediction accuracy on single-period sequences under a non-invariant transformation ($f(t+T)=2\cdot f(t)$) showing Transformer's failure with RoPE.}
    \vspace{-0.5cm}
    \label{fig:single_no_ac}
\end{figure}

\begin{figure}[ht]
    \centering
    \includegraphics[width=0.55\linewidth]{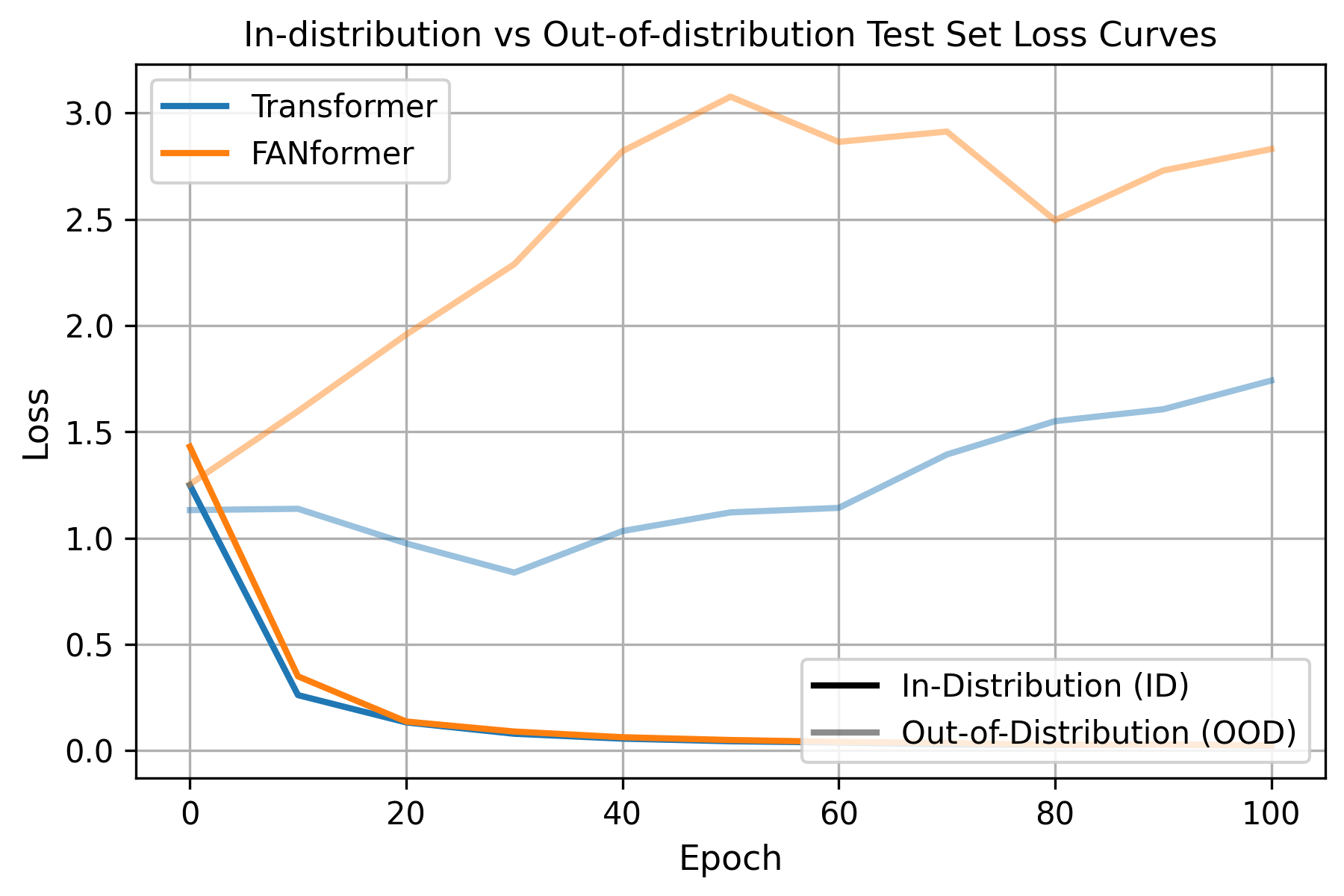}
    \caption{Loss curves for Transformer and FANformer with RoPE on single-period sequences under non-invariant transformations.}
    \vspace{-0.5cm}
    \label{fig:single_no_loss}
\end{figure}

This appendix presents a counterexample showing that RoPE fails to model single periodicity when the underlying transformation violates relative positional invariance.

Consider a single-period sequence continuation task with period length $T$. The periodic transformation is defined as
\[
f(t + T) = 2 \cdot f(t),
\]
which induces a scale change across periods. Under this transformation, the relative positional relation is no longer preserved:
\[
R_a(f(a)) - R_b(f(b)) \neq R_{a + T}(f(a + T)) - R_{b + T}(f(b + T)).
\]
Therefore, the core assumption required for RoPE-based periodic modeling does not hold.

We train a Transformer equipped with RoPE on this task using the same training protocol as in the successful single-periodicity experiment in Appendix~\ref{app:RoPE_SinPE}. The model fails to learn the periodic continuation rule. As shown in Fig.~\ref{fig:single_no_ac}, the prediction accuracy remains low across all evaluation settings. The training loss curve in Fig.~\ref{fig:single_no_loss} further indicates that the model does not converge to a stable solution.

These results demonstrate that RoPE cannot model single periodicity when the periodic structure involves non-invariant transformations. RoPE relies on relative positional consistency across cycles. When such consistency is violated, RoPE fails to support effective learning and generalization, even in the single-period setting.

This counterexample provides further verification of Eq.~\eqref{eq:R_no_equal}.
Regardless of how the equation becomes invalid, whether by adding a composite periodicity as in Section~\ref{subsubsec:rule_periodicity} or by scaling the original periodic sequence, the model consistently fails to generalize the periodicity. Hence, the relative position invariance instilled by RoPE is crucial for periodicity generalization.

\section{Group-Theoretic Interpretation of Circular Convolution}
\label{app:circular_convolution}

In Section~\ref{subsec:exp_composition}, we conduct an experiment utilizing circular convolution as another example of composite periodicity. Here we provide a theoretic explanation, demonstrating how circular convolution effectively represents composite periodicity.

Let $f_1, f_2 : \mathbb{Z}_N \to \mathbb{R}$ be two input sequences with period $P_1, P_2$, where $N$ is a common multiple of $P_1$ and $P_2$, usually $N = \operatorname{lcm}(P_1, P_2)$.
The circular convolution of $f_1, f_2$ is defined as
\begin{equation}
(f_1 * f_2)(t) := \sum_{n = 0}^{N - 1} f_1(n) \cdot f_2(t - n),
\label{eq:circular_convolution}
\end{equation}
where $n$ and $t - n$ are calculated under modulo $P_1$ and $P_2$ respectively.

Now we use the group $G = \langle g \rangle \simeq \mathbb{Z}_N$ to induce an action on the sequence space $\mathcal{F} = \{ f \mid f : \mathbb{Z}_N \to \mathbb{R} \}$.
The group action is defined as
\[
g \cdot t = t + 1 \pmod{N}.
\]
The induced action is\footnote{Here we use the inverse to ensure the associativity in the induced action. See footnote~\ref{fn:action}.}
\[
g \cdot f(t) = f(g^{-1} \cdot t) = f(t - 1 \bmod N) = f(t - 1).
\]
Hence, $g^n \cdot f(t) = f(g^{-n} \cdot t) = f(t - n)$.

For $i = 1, 2$, assume $H_i = \langle g^{P_i} \rangle$, since $f_i$ has a period of $P_i$, for all $h \in H_i$ we have $h \cdot f_i = f_i$, which implies the sequence periodicity. Because $P_i \mid N$, we have $g^n \cdot f_i(t) = f_i(t - n) = f_i(t - n \bmod{P_i})$.

Therefore, we can reform Eq.~\eqref{eq:circular_convolution} as:
\[
f_1 * f_2 = \sum_{n \in \mathbb{Z}_N} f_1(n) \cdot (g^n \cdot f_2).
\]

% Due to the commutativity of circular convolution, the group can act on $f_1$ either:
% \[
% f = f_1 * f_2 = \sum_{n \in \mathbb{Z}_N} (g^n \cdot f_1) \cdot f_2(n).
% \]

Next we prove that $g^k \cdot (f_1 * f_2) = (g^k \cdot f_1) * f_2$. On one hand,
\[
(g^k \cdot (f_1 * f_2))(t) = (f_1 * f_2) (t - k) = \sum_{n \in \mathbb{Z}_N} f_1(n) \cdot f_2(t - k - n).
\]
On the other hand,
\[
((g^k \cdot f_1) * f_2)(t) = \sum_{n \in \mathbb{Z}_N} (g^k \cdot f_1)(n) \cdot f_2(t - n) = \sum_{n \in \mathbb{Z}_N} f_1(n - k) \cdot f_2(t - n).
\]
Replace $n - k$ with $m$, then $n = k + m$, and as $n$ ranges over $\mathbb{Z}_N$, so does $m$. We have
\[
((g^k \cdot f_1) * f_2)(t) = \sum_{m \in \mathbb{Z}_N} f_1(m) \cdot f_2(t - k - m).
\]
This is the same formula as above, hence we have proved the equation.

Let $P = \operatorname{lcm}(P_1, P_2)$, then $g^P \cdot f_1 = f_1$, and $g^P \cdot f_2 = f_2$. From this we know that the circular convolution $f_1 * f_2$ has a period of $P$:
\[
g^P \cdot (f_1 * f_2) = (g^P \cdot f_1) * f_2 = f_1 * f_2, \quad i.e. \; (f_1 * f_2)(t - P) = (f_1 * f_2)(t).
\]
This is exactly the composite periodicity of two independent sequence periods.

%\section{Complete Results for Failure of Periodicity Generalization Across Models (Experiment 1)}
%\label{app:RQ1_full}

%The full heatmaps for all baseline models on composite periodicity tasks are shown in Fig.~\ref{fig:RQ1_full_heatmaps}.

\begin{figure}[t]
    \centering
    \begin{minipage}{0.48\linewidth}
        \centering
        \includegraphics[width=\linewidth]{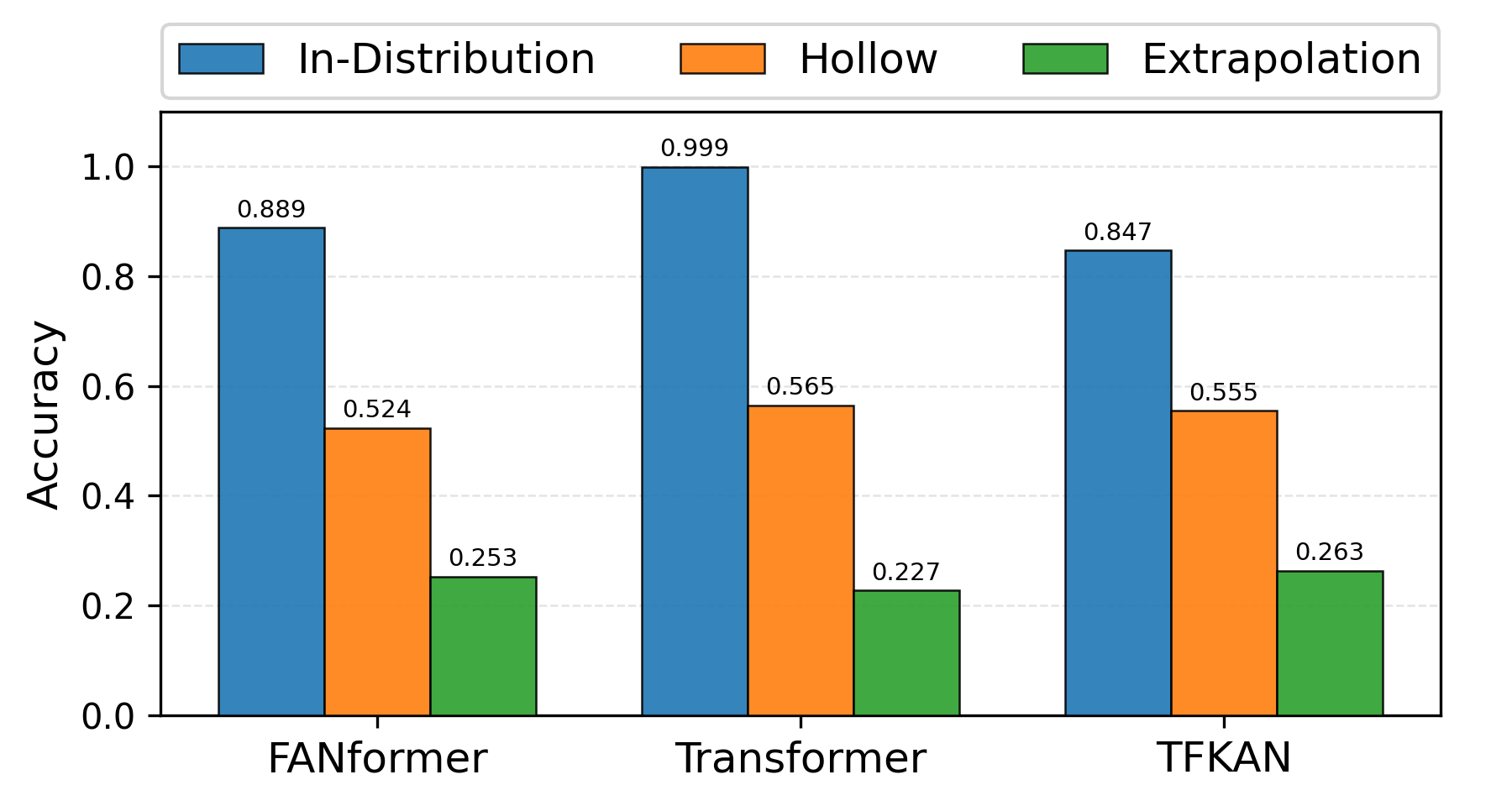}
        \captionof{figure}{Category accuracy heatmap across models after increasing data density.}
        \label{fig:rq2_category}
    \end{minipage}
    \hfill
    \begin{minipage}{0.48\linewidth}
        \centering
        \includegraphics[width=\linewidth]{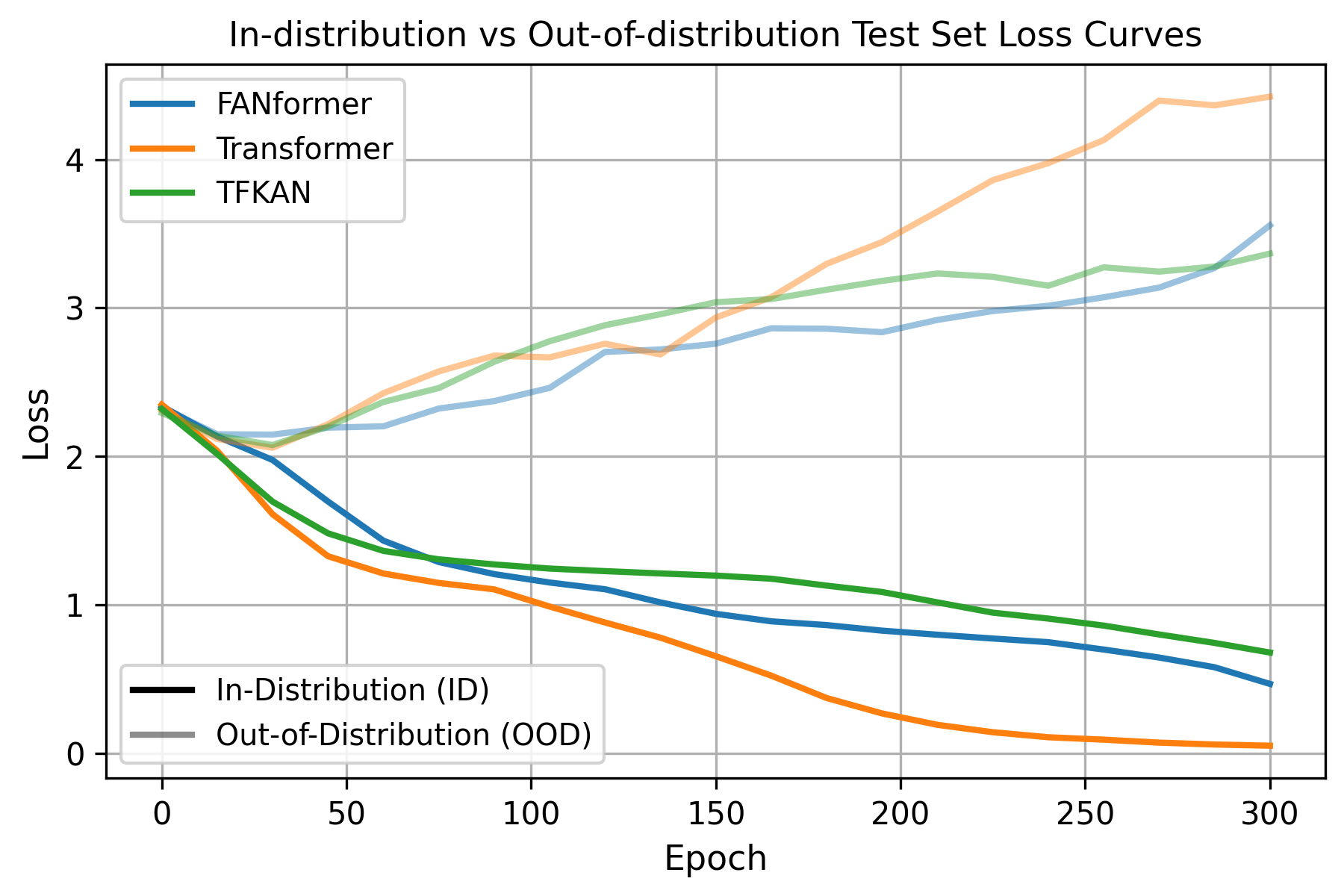}
        \captionof{figure}{Loss curves on ID and OOD test sets after increasing data density.}
        \label{fig:rq2_loss_curve}
    \end{minipage}
\end{figure}

\begin{table}[ht]
\centering
\caption{Effect of model scaling on composite periodicity generalization. Importantly, OOD generalization can only be interpreted when in-distribution fitting is sufficient. With different ID performance and loss, OOD loss and OOD accuracy are not strictly positively correlated.}

\setlength{\tabcolsep}{4pt}
\resizebox{0.57\linewidth}{!}{
\begin{tabular}{lccc|cccc}
\toprule
\multirow{2}{*}{Model}
& \multirow{2}{*}{Layers}
& \multicolumn{2}{c|}{Testset Loss $\downarrow$} 
& \multirow{2}{*}{ID} 
& \multirow{2}{*}{Hollow} 
& \multirow{2}{*}{Extrapolation} 
& \multirow{2}{*}{Avg.} \\
\cmidrule(lr){3-4}
& 
& ID 
& OOD
& 
& 
& 
& \\
\midrule
Transformer & 3 & 0.34 & 3.63 & 94.7 & 29.8 & 22.6 & 49.0 \\
Transformer & 5 & 0.05 & 3.89 & 99.8 & 36.4 & 24.4 & 53.5 \\
Transformer & 7 & 0.02 & 2.88 & 99.9 & 57.2 & 32.7 & 63.3 \\
\midrule
FANFormer   & 3 & 0.91 & 3.21 & 71.0 & 36.1 & 21.4 & 48.8 \\
FANFormer   & 5 & 0.47 & 3.15 & 89.5 & 41.7 & 19.4 & 50.2 \\
FANFormer   & 7 & 0.61 & 2.09 & 83.1 & 47.5 & 25.2 & 51.9 \\
\midrule
TFKAN         & 3 & 0.45 & 3.73 & 91.5 & 31.7 & 20.9 & 48.0 \\
TFKAN         & 5 & 0.17 & 2.38 & 98.1 & 50.5 & 27.6 & 58.7 \\
TFKAN         & 7 & 0.21 & 2.14 & 97.4 & 52.4 & 35.4 & 61.7 \\
\midrule
Mamba         & 3 & 1.36 & 1.42 & 49.4 & 48.4 & 46.7 & 48.2 \\
Mamba         & 5 & 1.33 & 1.48 & 49.5 & 43.1 & 45.3 & 46.0 \\
Mamba         & 7 & 1.47 & 1.63 & 45.6 & 42.0 & 36.0 & 41.2 \\
\midrule
RWKV         & 3 & 1.08 & 1.14 & 60.5 & 57.9 & 59.5 & 59.3 \\
RWKV         & 5 & 0.95 & 1.09 & 64.4 & 62.6 & 60.4 & 62.5  \\
RWKV         & 7 & 0.96 & 1.03 & 65.3 & 65.0 & 62.1 & 64.1 \\
\bottomrule
\end{tabular}
}
\label{tab:scaling_comparison}
\end{table}

\section{Complete Results for Effect of Training Data Density on OOD Generalization (Experiment 2)}
\label{app:RQ2_full}

The full category accuracy bar and the loss curves after increasing data density are shown in Fig.~\ref{fig:rq2_category} and Fig.~\ref{fig:rq2_loss_curve}. TFKAN~\cite{TFKAN} employs a dual time–frequency KAN~\cite{KAN} architecture that models nonlinear relationships and periodicity in time and frequency domains. In the following appendix, we include TFKAN as a baseline. The experimental results of KAN (or TFKAN) are similar to those reported in prior work~\cite{KAN_phd}. 

% The full results are shown in Fig.~\ref{fig:rq2_category} and Fig.~\ref{fig:rq2_loss_curve}.

\section{Complete Experimental Results for Scaling Model Parameters (Experiment 3)}
\label{app:RQ3_full}

The full results for different model and different parameter scales are shown in Table~\ref{tab:scaling_comparison}.

\section{Complete Experimental Results for the Circular Convolution Task (Experiment 4)}
\label{app:RQ4_full}
\begin{figure}[h]
    \centering
    \includegraphics[width=1.0\linewidth]{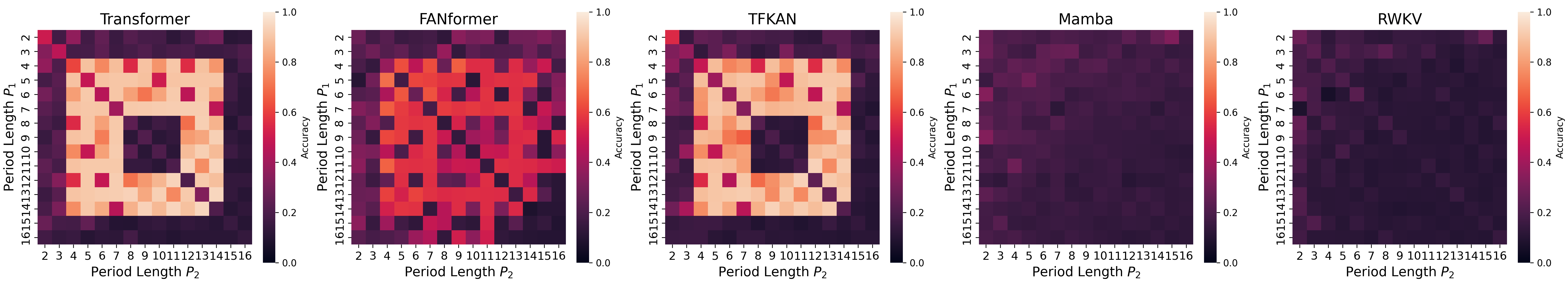}
    \caption{Heatmaps of model performance (450 training epochs) on the circular convolution task on different period settings.}
    \label{fig:RQ4_full_heatmaps}
\end{figure}

\begin{figure}[h]
    \centering
    \begin{minipage}{0.48\linewidth}
        \centering
        \includegraphics[width=\linewidth]{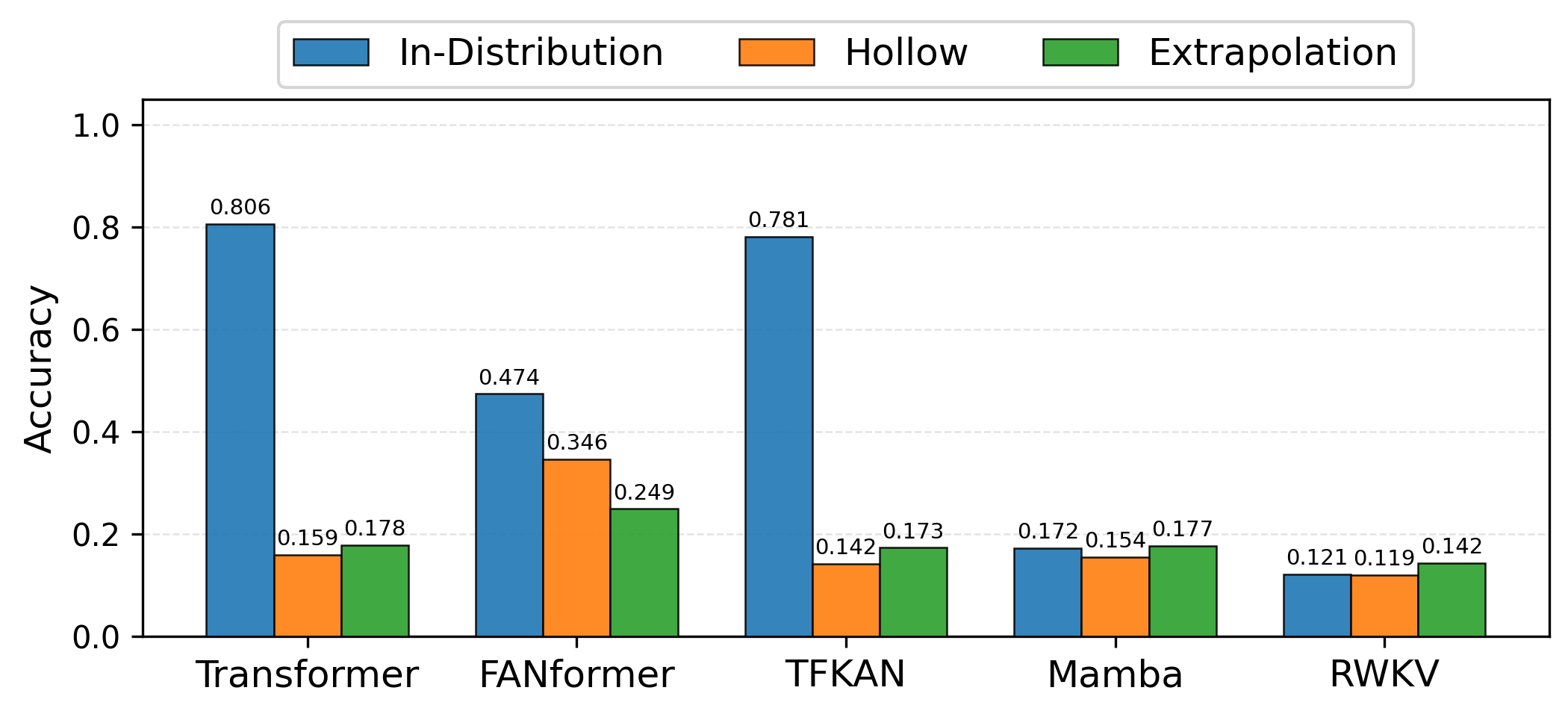}
        \captionof{figure}{Accuracy by category of different models on the circular convolution task on various OOD settings.}
        \label{fig:RQ4_category}
    \end{minipage}
    \hfill
    \begin{minipage}{0.48\linewidth}
        \centering
        \includegraphics[width=\linewidth]{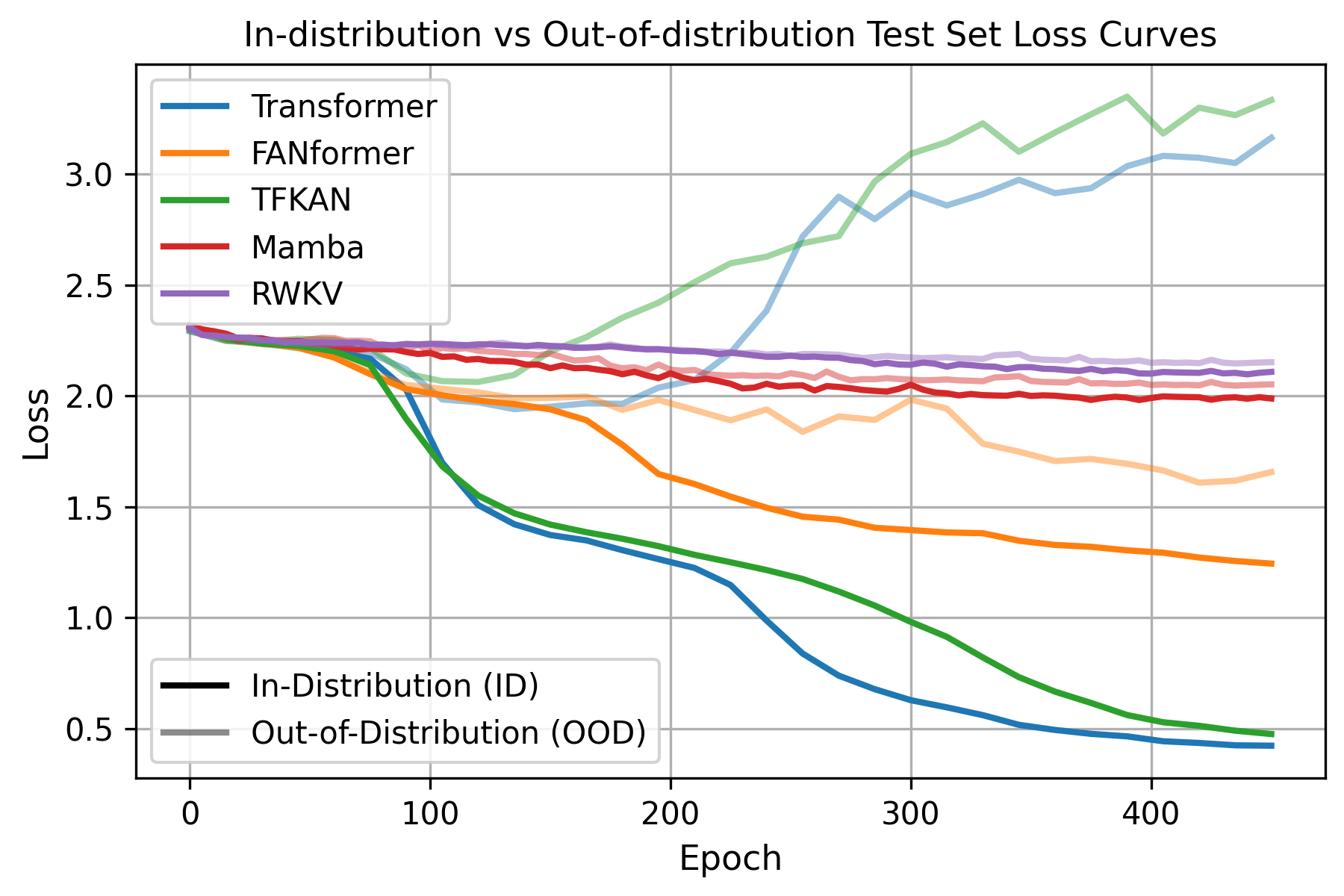}
        \captionof{figure}{Loss curves on ID and OOD test sets during training on circular convolution tasks.}
        \label{fig:RQ4_loss_curve}
    \end{minipage}
\end{figure}

The full accuracy heatmap, the category accuracy bar, and the loss curves for the circular convolution task are shown in Fig.~\ref{fig:RQ4_full_heatmaps}, Fig.~\ref{fig:RQ4_category}, and Fig.~\ref{fig:RQ4_loss_curve}, respectively.

% The full results are shown in Fig.~\ref{fig:RQ4_full_heatmaps}, Fig.~\ref{fig:RQ4_category} and Fig.~\ref{fig:RQ4_loss_curve}.

\section{Complete Experimental Results for More Composite Periodicity}
\label{app:more_composite}

In Section~\ref{sec:experiment}, our main experiment evaluates models' ability to generalize composite periodicity using the \textbf{Coper} dataset introduced in Section~\ref{sec:dataset}.
We utilize two rules: addition and modulo operation, which are both invariant, \ie both with period $T = 1$.
In this appendix, we further test models on additional composite periodicity to extend the concept of rule periodicity.
% Hence, the focus point in the main experiment is the composition of two different sequence periods.
% This appendix aims to further extend the concept of composite periodicity.
We replace the addition rule with a periodic operation: applying addition and subtraction alternately.
The composite operation can be formalized as
\[
\begin{aligned}
    C'(f_1(t), f_2(t)) & = (M \circ R')(f_1(t), f_2(t)) \\
    & = 
    \begin{cases}
        \bigl( f_1(t \bmod{P_1}) + f_2(t \bmod{P_2}) \bigr) \bmod{P}, \quad 2 \mid t \\
        \bigl( f_1(t \bmod{P_1}) - f_2(t \bmod{P_2}) \bigr) \bmod{P}, \quad 2 \nmid t
    \end{cases} \\
    & = \bigl( f_1(t \bmod{P_1}) + (-1)^t f_2(t \bmod{P_2}) \bigr) \bmod{P}.
\end{aligned}
\]
Consequently, the period of rule $R'$ becomes $T = 2$, and $R_{t + 2}' = R_{t}'$.

For example, let $f_1(t) = (1, 2, 3)$ with period $P_1 = 3$ and $f_2(t) = (1, 2)$ with period $P_2 = 2$. Then, their composition produces
\[
C'(f_1(t), f_2(t)) = \bigl( f_1(t \bmod{3}) + (-1)^t f_2(t \bmod{2}) \bigr) \bmod{P} = (2, 0, 4, 9, 3, 1, \dots).
\]

\begin{figure}[h]
    \centering
    \includegraphics[width=1.0\linewidth]{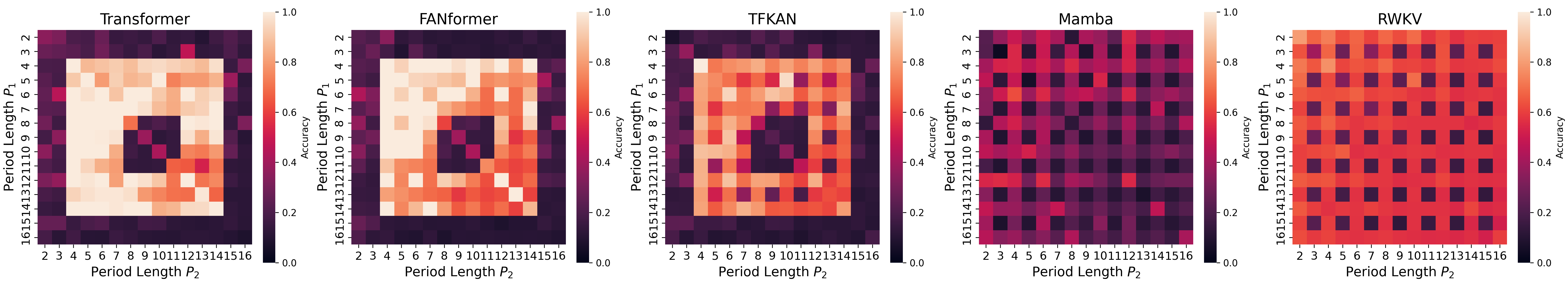}
    \caption{Heatmaps of model performance (450 training epochs) on the add–subtract alternating task on different period settings.}
    \label{fig:add_sub_heatmaps}
\end{figure}

\begin{figure}[t]
    \centering
    \begin{minipage}{0.48\linewidth}
        \centering
        \includegraphics[width=\linewidth]{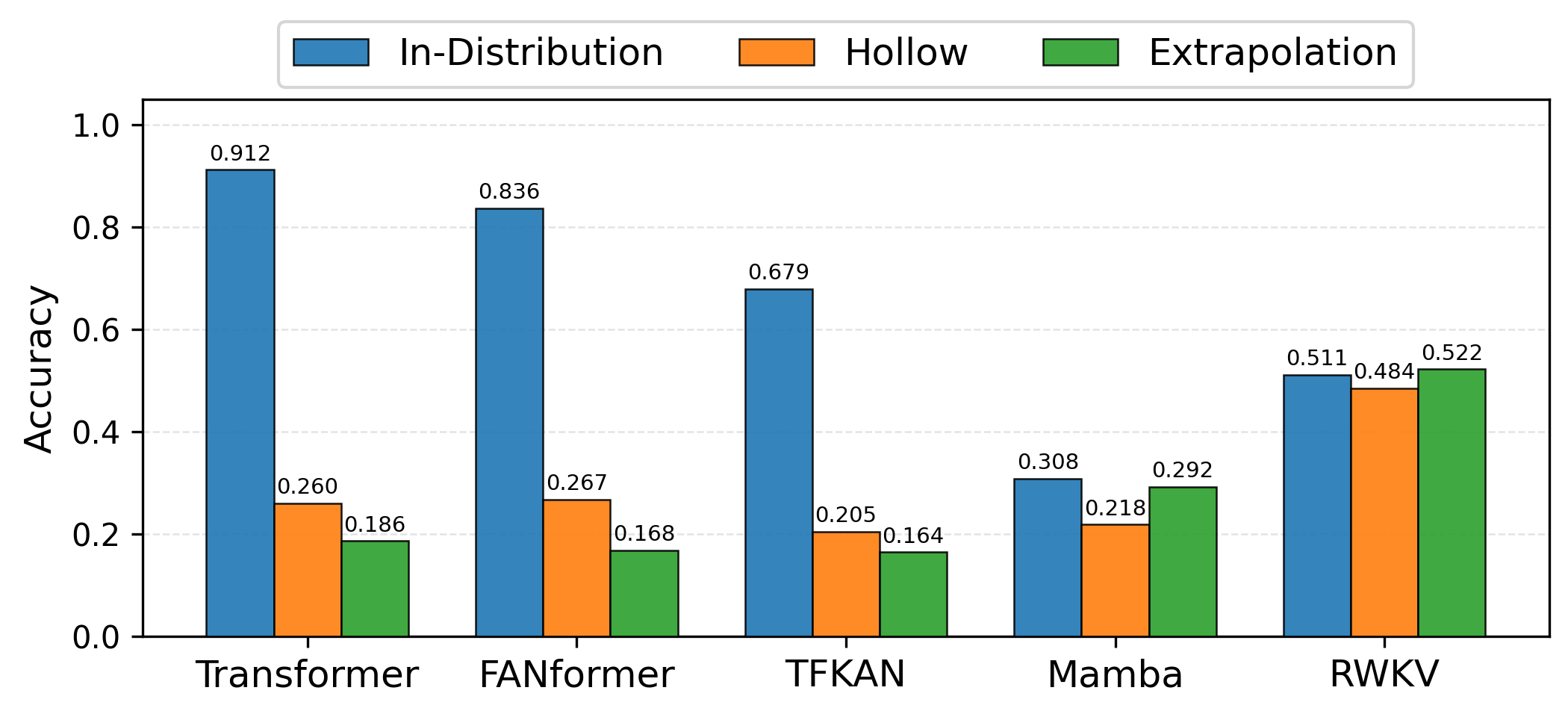}
        \captionof{figure}{Accuracy by category of different models on the add-subtract alternating task under various OOD settings.}
        \label{fig:add_sub_category}
    \end{minipage}
    \hfill
    \begin{minipage}{0.48\linewidth}
        \centering
        \includegraphics[width=\linewidth]{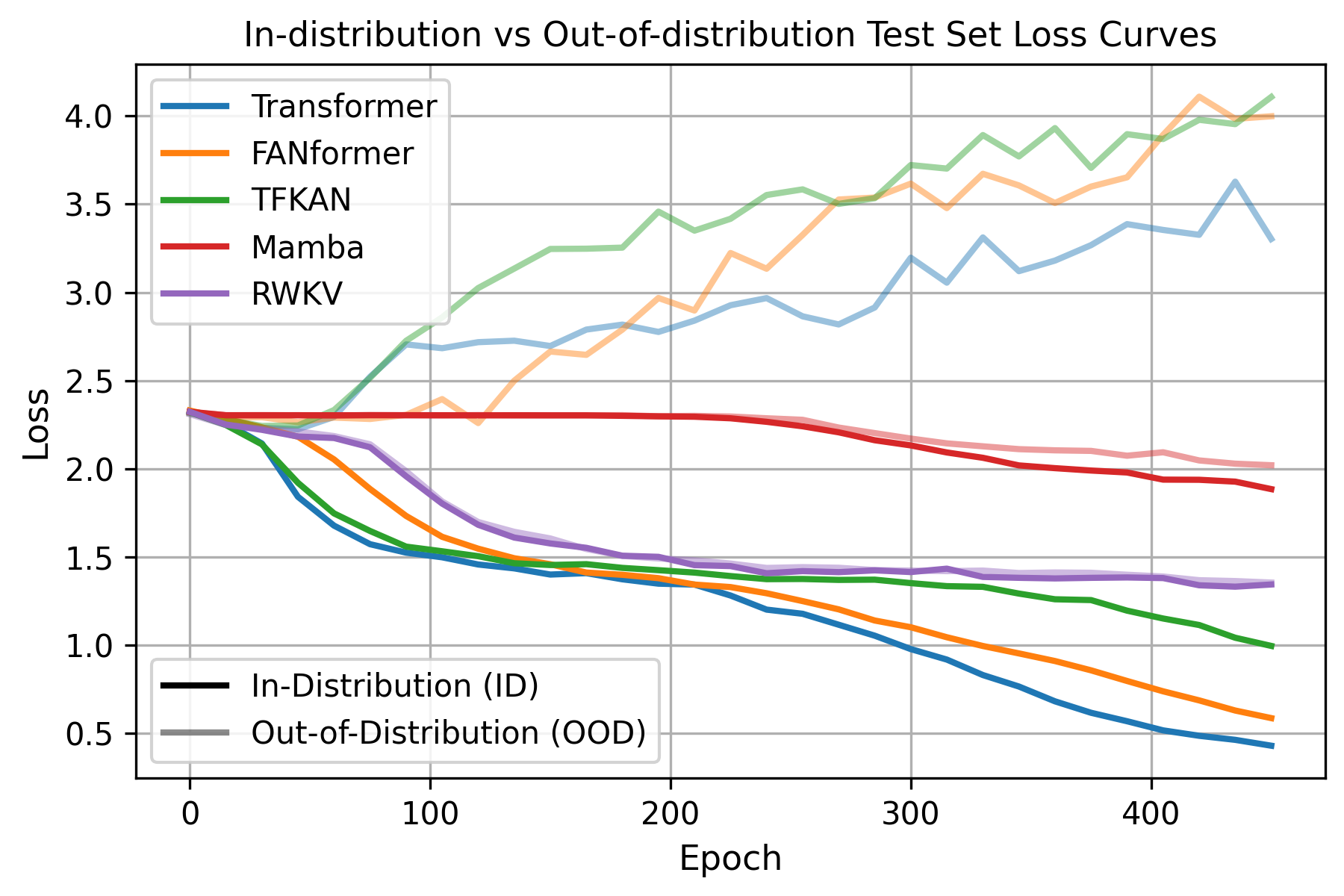}
        \captionof{figure}{Training loss curves of different models on the add-subtract alternating task.}
        \label{fig:add_sub_loss_curve}
    \end{minipage}
\end{figure}

Fig.~\ref{fig:add_sub_heatmaps}, Fig.~\ref{fig:add_sub_loss_curve}, and Fig.~\ref{fig:add_sub_category} show the results of the experiment. %, and the outcome is as predicted.
Just like what happens in Section~\ref{sec:experiment}, the models fail to generalize the periodic patterns, whether in the hollow or in the extrapolation area.
This experiment provides an additional verification of the conclusion in Section~\ref{subsubsec:rule_periodicity}: when it fails to retain relative position difference, Transformer with RoPE fails to represent such periodicity.

\section{Further Analysis of Hidden Periodicity}
\label{app:hidden_periodicity}

This Appendix evaluates the generalization ability of the Transformer and other periodicity modeling works, such as FANFormer, on hidden periodicity. The experiment trains models to fit $y = \sin(x)$. The training data is divided into in-distribution $[-3\pi, +3\pi]$ for training and OOD beyond this interval.  

\begin{figure}[h]
    \centering
    \includegraphics[width=1.0\linewidth]{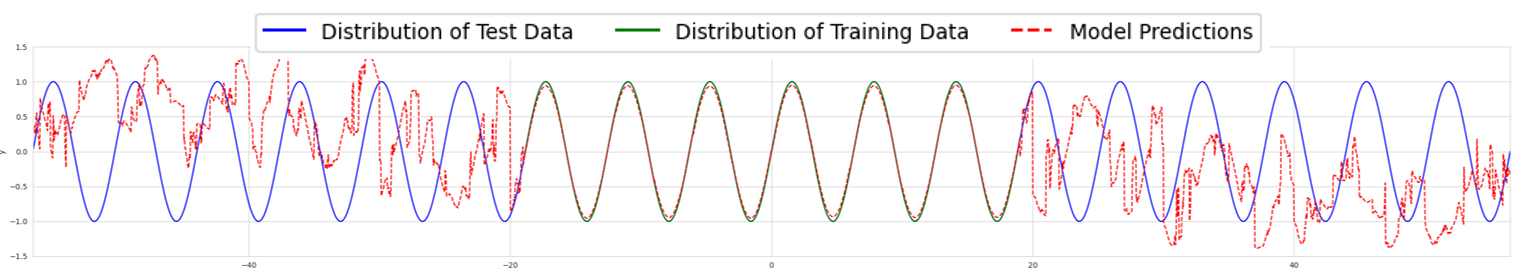}
    \caption{FANformer fails to generalize y = sin(x), where x is a token sequence}
    \label{fig:fan_sin}
\end{figure}

To enable the Transformer's self-attention mechanism, $x$ is tokenized into a fixed 10-digit sequence (\eg +3.1415926), including the sign ($+$ or $-$), the decimal point, and digits $0$--$9$. The model output is decoded back to numerical values using the same digit alignment. The Qwen2.5 tokenizer and embedding are frozen during training. As shown in Figure~\ref{fig:fan_sin}, FANFormer also fails to generalize hidden periodicity, similar to the Transformer. 

%%%%%%%%%%%%%%%%%%%%%%%%%%%%%%%%%%%%%%%%%%%%%%%%%%%%%%%%%%%%%%%%%%%%%%%%%%%%%%%
%%%%%%%%%%%%%%%%%%%%%%%%%%%%%%%%%%%%%%%%%%%%%%%%%%%%%%%%%%%%%%%%%%%%%%%%%%%%%%%

\end{document}